\documentclass[sigconf, nonacm]{acmart}

\newcommand\vldbdoi{XX.XX/XXX.XX}

\newcommand\vldbpagestyle{plain} 

\usepackage{graphicx}
\usepackage{subcaption}
\usepackage{listings}
\usepackage{enumitem}
\usepackage{booktabs}
\usepackage{pifont}
\usepackage{makecell}
\usepackage[most]{tcolorbox}

\usepackage{microtype}
\usepackage{multirow}
\usepackage{threeparttable}

\newcommand{\blue}[1]{\textcolor{black}{#1}}

\newcommand{\minihead}[1]{{\vspace{.55em}\noindent\textbf{#1.} }}
\newcommand{\miniheadnodot}[1]{{\vspace{.55em}\noindent\textbf{#1} }}

\newcommand{\mycross}{\textcolor{red}{\textbf{$\times$}}}
\newcommand{\mycheck}{\textcolor{ForestGreen}{\textbf{\checkmark}}}

\newcommand{\syntaxhighlight}[1]{\textbf{\textcolor{blue}{#1}}}
\newcolumntype{C}[1]{>{\centering\arraybackslash}m{#1}}

\newcommand*\LSTfont{\Small\ttfamily\SetTracking{encoding=*}{-60}\lsstyle}

\lstdefinestyle{yaml}{
    frame=single,
    language=,
    basicstyle=\LSTfont, 
    keywordstyle={[2]\color{blue}\bfseries}, 
    sensitive=true,
    literate=
        {name:}{{\color{blue}\bfseries name:}}1
        {description:}{{\color{blue}\bfseries description:}}1
        {columns:}{{\color{blue}\bfseries columns:}}1
        {models:}{{\color{blue}\bfseries models:}}1
}

\lstdefinestyle{yaml}{
    frame=single,
    language=,
    basicstyle=\LSTfont,
    morekeywords={[2]{name, description, columns, models}},
    keywordstyle={[2]\color{blue}},
    sensitive=true,
    literate=
      {:}{{{\color{blue}{:}}}}1
}

\lstdefinestyle{yamll}{
    frame=single,
    language=,
    basicstyle=\LSTfont,
    sensitive=true,
    morekeywords={[3]Airbyte, config, files_definition_id, workspace_id, flat_files, format, path, sync_mode, table, snowflake, account, database, password, role, schema, username, warehouse, connection_string, mongodb}, 
    morekeywords={[2]'mongodb,elt-mongodb}, 
    keywordstyle={[3]\color{blue}},   
    keywordstyle={[2]\color{black}},  
    alsoletter={'-}
}

\lstdefinestyle{terraform}{
    frame=single,
    language=,
    basicstyle=\LSTfont,
    morekeywords={[2]{resource, name, definition_id , workspace_id, configuration, provider, https_public_web, url, format, dataset_name, source_id, destination_id, namespace_definition, configurations, streams, sync_mode, database_config, self_managed_replica_set, connection_string }},
    morekeywords={[3].destination_id,.source_id}, 
    keywordstyle={[2]\color{blue}},
    keywordstyle={[3]\color{black}},
    sensitive=true,
    alsoletter={.}
}

\lstdefinestyle{dbt}{
    frame=single,
    language=,
    basicstyle=\LSTfont,
    morekeywords={[2]{my_dbt_profile, target , outputs , dev , type, account, https_public_web, user, password, role, database, warehouse, schema}},
    keywordstyle={[2]\color{blue}},
    sensitive=true,
    literate=
      {:}{{{\color{blue}{:}}}}1
}

\usepackage{threeparttable}
\usepackage{balance}

\newcommand{\agent}{SAR-Agent }

\AtBeginDocument{%
  }

\makeatletter
\def\@copyrightspace{\relax}
\makeatother

\begin{document}

\title{
Pervasive Annotation Errors Break Text-to-SQL Benchmarks and Leaderboards}

\author{Tengjun Jin}
\affiliation{%
  \institution{University of Illinois (UIUC)}
  \city{Urbana}
  \country{USA}
}
\email{tengjun2@illinois.edu}

\author{Yoojin Choi}
\affiliation{%
  \institution{University of Illinois (UIUC)}
  \city{Urbana}
  \country{USA}
}
\email{yoojinc3@illinois.edu}

\author{Yuxuan Zhu}
\affiliation{%
  \institution{University of Illinois (UIUC)}
  \city{Urbana}
  \country{USA}
}
\email{yxx404@illinois.edu}

\author{Daniel Kang}
\affiliation{%
  \institution{University of Illinois (UIUC)}
  \city{Urbana}
  \country{USA}
}
\email{ddkang@illinois.edu}


\begin{abstract}
Researchers have proposed numerous text-to-SQL techniques to streamline data analytics and accelerate the development of data-driven applications.
To compare these techniques and select the best one for deployment, the community depends on public benchmarks and their leaderboards. Since these benchmarks heavily rely on human annotations during question construction and answer evaluation, the validity of the annotations is crucial.


 \blue{In this paper, we conduct an empirical study that (i) benchmarks annotation error rates for two widely used text-to-SQL benchmarks, BIRD and Spider 2.0-Snow, and (ii) corrects a subset of the BIRD development (Dev) set to measure the impact of annotation errors on text-to-SQL agent performance and leaderboard rankings.
Through expert analysis, we show that BIRD Mini-Dev and Spider 2.0-Snow have error rates of 52.8\% and 62.8\%, respectively. We re-evaluate all 16 open-source agents from the BIRD leaderboard on both the original and the corrected BIRD Dev subsets.} We show that performance changes range from $-7$\% to $31$\% (in relative terms) and rank changes range from $-9$ to $+9$ positions. We further assess whether these impacts generalize to the full BIRD Dev set. We find that the rankings of  agents on the uncorrected subset correlate strongly with those on the full Dev set (Spearman's $r_s$=0.85, $p$=3.26e-5), whereas they correlate weakly with those on the corrected subset (Spearman's $r_s$=0.32, $p$=0.23). These findings show that annotation errors can significantly distort reported performance and rankings, potentially misguiding research directions or deployment choices. Our code and data are available at \url{https://github.com/uiuc-kang-lab/text_to_sql_benchmarks}.

\end{abstract}

\maketitle

\pagestyle{\vldbpagestyle}


\section{Introduction}
\label{sec:introduction}

As text-to-SQL techniques become increasingly important for data analytics and data-driven applications~\cite{XiYan, pourreza2023dinsqldecomposedincontextlearning, pourreza2024chasesqlmultipathreasoningpreference, pourreza2025reasoningsqlreinforcementlearningsql, opensearchsql, gao2023texttosqlempoweredlargelanguage, chess}, researchers and practitioners have introduced a variety of text-to-SQL benchmarks~\cite{IMDB, scholar,  ATIS, lee2021kaggledbqarealisticevaluationtexttosql, spider1, bird, wikisql, spider2, BEAVER, swesql}. These benchmarks guide researchers in improving text-to-SQL techniques and help practitioners select the best text-to-SQL technique for downstream applications in real-world scenarios.

Unfortunately, current text-to-SQL benchmarks are unreliable due to annotation errors. 
As reported in prior work, BIRD~\cite{bird} has annotation errors in 32\% of examples in the mini development set (Mini-Dev)~\cite{BIRDminidev} and 49\% of financial domain examples~\cite{wretblad2024understandingeffectsnoisetexttosql}. Such errors can distort reported performance and rankings of agents and mislead researchers and practitioners.

\blue{To understand how we can build reliable text-to-SQL benchmarks with minimal annotation errors, we conduct an in-depth analysis of annotation errors in text-to-SQL benchmarks.
We design our analysis based on four recurring text-to-SQL annotation error patterns.
Given a target database $\mathcal{D}$, human annotators curate text-to-SQL examples, each comprising a natural language input $\mathcal{T}$ and a ground-truth SQL query $\mathcal{Q}$.\footnote{Input $\mathcal{T}$ consists of a user question and, optionally, external knowledge.}} Annotation errors can arise within any individual component or from inconsistencies across components, \blue{including}:
\begin{enumerate}[label=\textbf{E\arabic*.},leftmargin=*]
\item Mismatches between the semantics of $\mathcal{Q}$ and the logic of $\mathcal{T}$.
\item Mismatches between the semantics of $\mathcal{Q}$ and $\mathcal{D}$ due to a limited understanding of the data or the schema.
\item Mismatches between the semantics of $\mathcal{Q}$ and the domain knowledge relevant to $\mathcal{T}$, \blue{or misannotated domain knowledge in $\mathcal{T}$}.
\item Ambiguity in $\mathcal{T}$, such as multiple possible interpretations or an unclear output format.
\end{enumerate}
Figure~\ref{sub:spider2_issue} illustrates a misuse of a Snowflake function (\textbf{E1}). Figure~\ref{sub:bird_issue} presents an incorrect annotation from BIRD due to the annotator's lack of domain knowledge (\textbf{E3}). We present additional representative examples for each error pattern in Section~\ref{subsec:anno_issue}.

\begin{figure}[t]
  \begin{subfigure}[t]{\columnwidth}
    \input{figures/sf_bq291}
\caption{sf\_bq291 of Spider 2.0-Snow. Annotators mistakenly swapped the longitude–latitude order in the \texttt{ST\_POINT} arguments \textbf{(E1)}.}
\label{sub:spider2_issue}
\end{subfigure}

 \vspace{1em}

  \begin{subfigure}[t]{\columnwidth}
    \input{figures/bird46}
\caption{Example 46 of BIRD Mini-Dev. Annotators had limited understanding of ``K-12'' \textbf{(E3)}.}
\label{sub:bird_issue}
\end{subfigure}





\caption{Two examples from existing text-to-SQL benchmarks that demonstrate incorrect annotations.}
\label{fig:issued_sql}
\end{figure}

\blue{We conducted a human-in-the-loop, three-stage audit, with all final decisions made by human SQL experts, to benchmark annotation error rates for two widely used benchmarks, BIRD ~\cite{bird} and Spider 2.0-Snow~\cite{spider2}. To facilitate the audit, we developed SAR-Agent (\textbf{S}QL \textbf{A}nnotation \textbf{R}eviewer agent), the first AI agent  that assists SQL experts in detecting annotation errors in text-to-SQL benchmarks. Our three-stage audit consisted of: (1) SAR-Agent producing per-annotation diagnostic reports; (2) human SQL experts adjudicating all agent-flagged cases, verifying the stated error reasons and assigning final correctness labels; and (3) human SQL experts extending manual review to additional examples based on the recurring error patterns.}

\blue{We manually corrected a subset of the BIRD Dev set to measure the impact of annotation errors on text-to-SQL agent performance and leaderboard rankings. To support manual corrections by SQL experts, we introduced SAPAR (\textbf{S}QL \textbf{A}nnotation \textbf{P}ipeline with an AI \textbf{A}gent \textbf{R}eviewer),  which integrated SAR-Agent into the standard text-to-SQL annotation pipeline \cite{bird}. We randomly sampled 100 of the 1,534 BIRD Dev examples and, guided by SAPAR, manually corrected all identified errors. We then re-evaluated all 16 open-source text-to-SQL agents from the BIRD leaderboard on both the original and the corrected Dev subsets.}




\blue{
Below, we (i) quantify annotation error rates in BIRD Mini-Dev (an official subset of the BIRD Dev set) \cite{bird} and Spider 2.0-Snow \cite{spider2}, (ii) analyze their impact on agent performance and leaderboard rankings, and (iii) evaluate SAR-Agent’s error-detection capability to demonstrate its utility for improving benchmark quality.}


\minihead{High error rates in existing text-to-SQL benchmarks} Our analysis reveals a wider range of annotation errors in text-to-SQL benchmarks than previously recognized. We executed \agent on BIRD Mini-Dev and Spider 2.0-Snow to generate diagnostic reports for each example. After manual verification, we find that 52.8\% of the examples in BIRD Mini-Dev~\cite{bird} contain annotation errors, which is 20.5\% higher than the previously reported rate of 32.3\%~\cite{BIRDminidev}. In addition, we are the first to comprehensively examine annotation errors in Spider 2.0-Snow ~\cite{spider2}, a recently released benchmark and successor to the widely used Spider 1.0 ~\cite{spider1}. The Spider 2.0 team applied large language models (LLMs) to paraphrase user questions for clarity and corrected 45\% of examples in the first validation round and 5\% in the second~\cite{spider2}. Nevertheless, across all 121 problems for which ground-truth SQL is publicly available, we still identify an annotation error rate of 62.8\%.

\minihead{The unreliability of the BIRD leaderboard}
These annotation errors cause severe misestimation and misranking of agents' performance. \blue{Through our re-evaluation of 16 agents on both the original and the corrected BIRD Dev subsets,} we identify significant performance changes ranging from $-7$\% to $31$\% in relative terms and ranking position changes ranging from $-9$ to $+9$. For instance, we find the performance of CHESS~\cite{chess}, an agent previously ranked 7th among our 16 selected agents, increases from 62\% to 81\% after corrections and moves to 1st place.

We further computed two Spearman's rank correlation coefficients ($r_s$) to assess the impact of annotation errors on agent rankings. We find that the rankings on the original Dev subset correlate strongly with those on the original full Dev set ($r_s$=0.85, $p$=3.26e-5), but have a weak correlation with the rankings on the corrected Dev subset ($r_s$=0.32, $p$= 0.23). 

These results demonstrate that annotation errors can distort reported performance and rankings, undermining the reliability of the leaderboard. Such unreliability can potentially misguide researchers and practitioners in improving text-to-SQL techniques and picking the best agent.


\minihead{The effectiveness of SAR-Agent} SAR-Agent demonstrates its effectiveness in detecting annotation errors. Our manual verification of diagnostic reports for each annotated example shows that SAR-Agent achieves 83\% precision on BIRD Mini-Dev. On Spider 2.0-Snow, for which no prior analyses exist, SAR-Agent achieves 89\% precision. We further compared SAR-Agent's detected error examples (after manual verification) on BIRD Mini-Dev with those reported by the Arcwise team's SQL experts~\cite{BIRDminidev}. Among the reported 161 error examples, SAR-Agent attains an 85.7\% hit rate. SAR-Agent also identifies 41.6\% more error examples than Arcwise.



\vspace{0.5em}

We summarize our contribution as follows:
\begin{enumerate}[leftmargin=*]
    \item \blue{Through an in-depth manual review by SQL experts, we find annotation error rates of 52.8\% in BIRD Mini-Dev and 62.8\% in Spider 2.0-Snow.}
    \item By re-evaluating all 16 open-source text-to-SQL agents from the BIRD leaderboard, we identify substantial performance shifts, with relative performance changes ranging from $-7$\% to $31$\% and ranking changes ranging from $-9$ to $+9$.
    \item \blue{We introduce a toolkit, comprising SAR-Agent and SAPAR, for annotation error detection and correction.}

\end{enumerate}

\begin{figure}[t]
    \centering
    \includegraphics[width=0.9\linewidth]{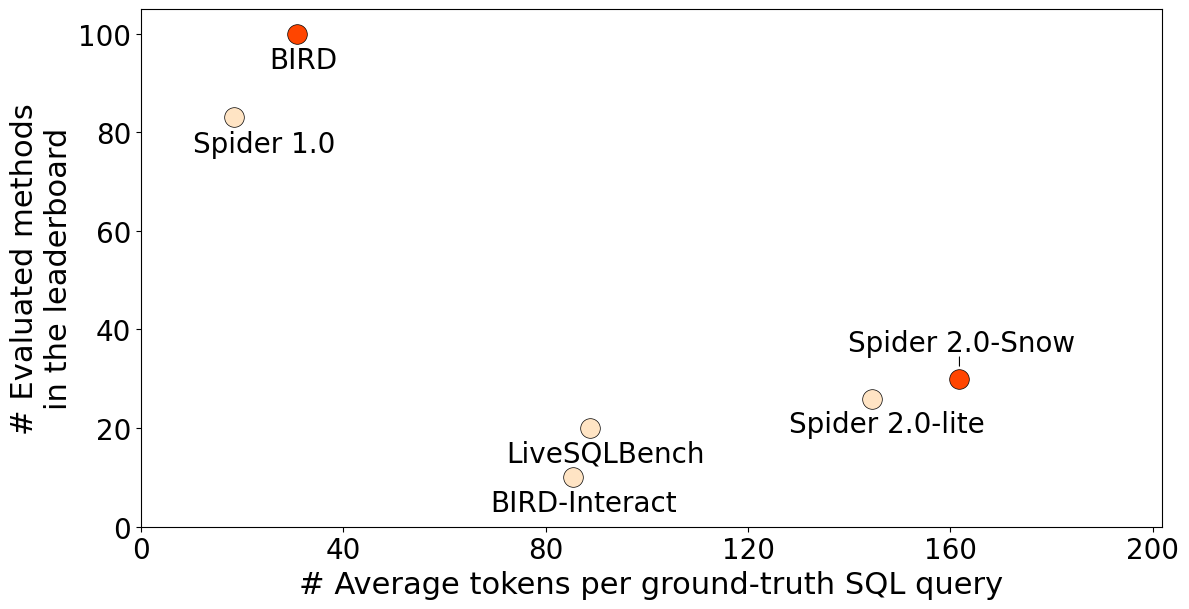}
    \caption{\blue{Number of evaluated methods in the leaderboard and ground-truth SQL tokens across text-to-SQL benchmarks (data cutoff: October 20, 2025).}}
    \label{fig:bench}
\end{figure}
\section{EXPERIMENT DESIGN}
In this work, we develop a toolkit to help human SQL experts analyze and correct existing text-to-SQL benchmarks and generate insights from them. We design experiments to investigate the following research questions:
\begin{enumerate}[label=\textbf{Q\arabic*.},leftmargin=*]
\item What are the annotation error rates in widely used text-to-SQL benchmarks?
\item How do annotation errors affect text-to-SQL agents' performance and rankings on the leaderboard?
\item How effective is SAR-Agent at detecting annotation errors?
\end{enumerate}

\subsection{Benchmark Selection}
\blue{We selected two text-to-SQL benchmarks, BIRD \cite{bird} and Spider 2.0-Snow \cite{spider2}, for our empirical study based on usage and query complexity.\footnote{BIRD note: \blue{BIRD consists of Train, Dev and Test sets. We focus on the Dev set because (i) the Train set is only used in studies that fine-tune LLMs, (ii) the Test set is hidden, and (iii) researchers rely on the Dev set for local evaluation and error analysis.}\\
Spider 2.0-Snow note: The Spider 2.0 gold queries were updated on Oct. 29, 2025. Since our evaluation was conducted before that, we used the annotations as of Aug. 20, 2025, available at: \url{https://github.com/xlang-ai/Spider2/tree/84911f375c87903c73ed5fe93769b593a14efb42/spider2-snow/evaluation_suite/gold/sql}} We quantified usage by the number of evaluated methods on each benchmark’s leaderboard, and complexity by the average number of tokens in the ground-truth SQL queries (using whitespace tokenization \cite{spider2}) across six benchmarks \cite{bird, spider1, spider2, birdinteract2025, livesqlbench2025}. As shown in Figure~\ref{fig:bench}, BIRD has the largest number (100) of evaluated methods in its leaderboard. Spider 2.0-Snow has the most complex SQL queries (highest average token count of its ground-truth SQL queries, 161.8 tokens per SQL query) and is the successor to the widely used Spider 1.0 \cite{spider1}. }

\subsection{Toolkit for Annotation Error Detection and Correction}
\blue{We developed a toolkit comprising SAR-Agent and SAPAR to facilitate manual analysis and correction of text-to-SQL annotation errors. For error detection, we ran SAR-Agent on the annotations to generate per-example diagnostic reports. SQL experts then reviewed the reported issues and assigned final correctness labels. For annotation error correction, we manually fixed the identified errors in a sampled BIRD Dev subset based on SAPAR. We provide implementation details of SAR-Agent and SAPAR in Section~\ref{sec:toolkit}.}

\subsection{Evaluation Overview}
We structure our experiments around our research questions.
In Section~\ref{sec:error_dectection}, we use a three-step examination procedure to detect annotation errors in two widely used benchmarks, BIRD Mini-Dev \cite{bird} and Spider 2.0-Snow \cite{spider2}, and quantify their error rates to address \textbf{Q1}.
 In Section~\ref{sec:reevaluate}, we sample 100 examples from the BIRD Dev set and manually correct annotation errors by applying SAPAR. To answer \textbf{Q2}, we then re-evaluate all 16 open-source agents listed on the BIRD leaderboard on both the original and the corrected versions of this sampled subset. We quantify changes in execution accuracy and the relative ranking among these agents. \blue{In Section~\ref{sec:SAR}, we address \textbf{Q3} by evaluating SAR-Agent’s performance and comparing the number of errors it detects with those identified in a prior audit by SQL experts from Arcwise \cite{BIRDminidev}.}


\begin{figure*}[t]
    \centering
    \includegraphics[width=0.9\linewidth]{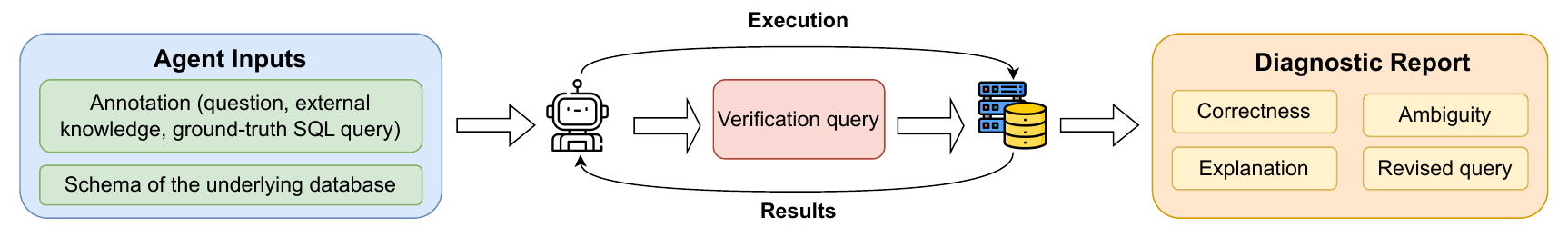}
    \caption{Overview of the SAR-Agent framework. At each iteration, the agent proposes and executes a query to validate the annotation. In the final iteration, it generates a diagnostic report that assesses the annotation's correctness and ambiguity.}
    \label{fig:agent_frame}
\end{figure*}

\begin{figure}[t]
    \centering
    \includegraphics[width=0.95\linewidth]{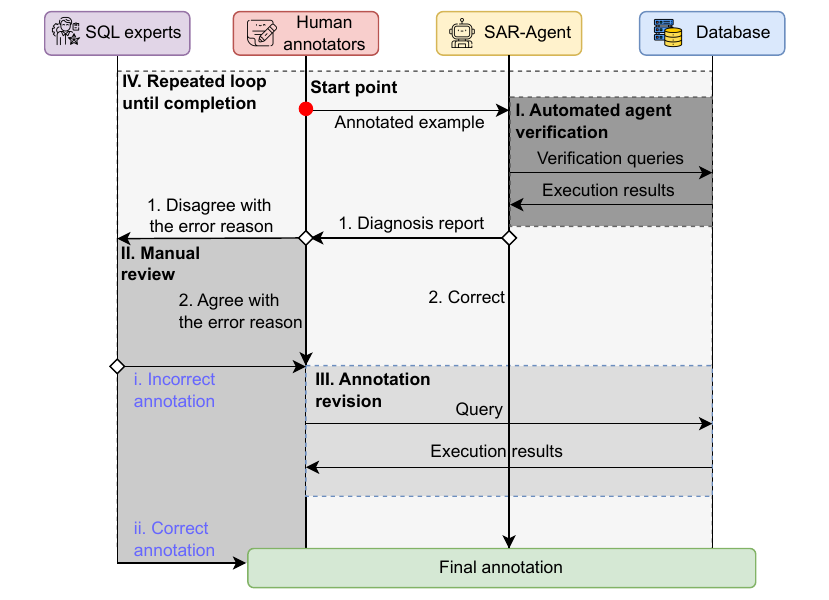}
    \caption{\blue{Overview of SAPAR. SAR-Agent audits the annotation and generates a diagnostic report for expert review. If errors are found, annotators iteratively revise the annotation until SAR-Agent or human SQL experts judge it correct.}}
    \label{fig:agent_pipelin}
\end{figure}

\section{Toolkit}
\label{sec:toolkit}
\blue{In this section, we describe the implementation of our toolkit for detecting and correcting annotation errors. We first present the design of SAR-Agent, then detail the SAPAR architecture.}


\subsection{SAR-Agent}
\blue{To support effective detection of annotation errors and assess error rates across different text-to-SQL benchmarks,  we develop SAR-Agent, a SQL Annotation Reviewer agent. SAR-Agent evaluates both the correctness and potential ambiguity of annotations.} In practice, data analysts typically explore databases and construct SQL queries in incremental steps, a process known as exploratory data analysis (EDA) \cite{eda, liu2025skyrlsql}. \blue{SAR-Agent mirrors this workflow by conducting multi-turn interactions with the database and verifying annotations step by step. We present an example demonstrating how SAR-Agent verifies annotations and generates diagnostic reports in Appendix \ref{subsec:end_example}.}

We present the framework of SAR-Agent in Figure~\ref{fig:agent_frame} and describe implementation details below.


\minihead{Core mechanism} \blue{Rather than relying on a single-turn LLM judgment, SAR-Agent verifies each annotation incrementally. In each iteration, it formulates a targeted verification query to probe a specific aspect of the annotated SQL, executes it against the database, and inspects the result. All prior verification queries and their execution results are retained in the agent’s memory to guide the next iteration. In the final step, the agent synthesizes its findings into a comprehensive diagnostic report.}

\minihead{Agent inputs} \blue{\agent takes as input the following:}
\begin{enumerate}[leftmargin=*]

\item \blue{\textit{Text-to-SQL annotation:} The natural language input $\mathcal{T}$ and the corresponding ground-truth SQL query $\mathcal{Q}$, both annotated by humans. $\mathcal{T}$ comprises the user question and, when applicable, external knowledge that supplements the question.}

\item \blue{\textit{Schema of the underlying database $\mathcal{D}$:} Table and column specifications, including table names, column names, and column descriptions. Given the schema complexity of Spider 2.0-Snow (an average of 812.1 columns per database \cite{spider2}), we restrict the schema to the gold tables  (i.e., those referenced by the ground-truth SQL query), as provided by the Spider team.\footnote{The gold tables for each query can be found at \url{https://github.com/xlang-ai/Spider2/blob/main/methods/gold-tables/spider2-snow-gold-tables.jsonl}.}}


\end{enumerate}

\minihead{Agent environment}
The environment includes the databases for each example. To verify annotations, the agent can directly access these databases by calling predefined functions.

\minihead{Agent outputs} SAR-Agent generates two files: one records the proposed verification queries, and the other presents the final diagnostic report. We describe these files as follows:
\begin{enumerate}[leftmargin=*]
\item \textit{Verification queries:} The agent logs each verification query it proposes, with accompanying explanations and the corresponding execution results.
\item \textit{Final diagnostic report:} After executing verification queries, the agent generates a comprehensive diagnostic report. This report includes a judgment on question ambiguity, an assessment of SQL query correctness, detailed explanations, and a proposed corrected query if an error is detected.\footnote{The corrected query is intended to help annotators understand the agent’s reasoning. We do not evaluate the accuracy of these corrected queries, as some examples require revisions to both the natural language question and the SQL query.}
\end{enumerate}

\minihead{Agent functions} SAR-Agent is implemented using a function calling framework.\footnote{We manually define the functions, and we can also equivalently implement the agent using Model Context Protocol (MCP).} We design functions that allow it to interact with the database and generate the final diagnostic report, including:
\begin{enumerate}[leftmargin=*]
    \item \texttt{execute\_query:} executes a query on the database and returns the execution results to the agent.
    \item \texttt{terminate:} generates a final diagnostic report and terminates.
\end{enumerate}

\subsection{SAPAR}
\blue{To enhance the efficiency and quality of annotation error correction, we propose SAPAR, which integrates SAR-Agent into the standard text-to-SQL annotation pipeline \cite{bird}. Existing pipelines require substantial human effort to ensure benchmark quality and provide rigorous evaluation. For example, BIRD \cite{bird} uses double-blind annotation followed by expert review. However, we still find pervasive annotation errors in BIRD Mini-Dev.}

SAPAR augments expert review with SAR-Agent, improving both the efficiency and accuracy of annotation. First, annotators can receive annotation diagnoses within minutes, enabling immediate corrections after initial annotation rather than at the end of a batch. Second, this workflow further increases efficiency by reducing the manual review effort of SQL experts. Text-to-SQL benchmarks are typically cross-domain (e.g., BIRD contains 37 domains), making it time-consuming for reviewers to master all necessary domain knowledge. SAR-Agent can reduce human efforts by eliminating the routine need for a SQL expert to review each annotation. Third, through our evaluation results in Section~\ref {sec:SAR}, we find that SAR-Agent can detect more errors than the human experts in prior work \cite{bird}, thereby improving annotation quality.

We show the overall architecture of SAPAR in Figure~\ref{fig:agent_pipelin}, and describe the workflow as follows:

    
    

\minihead{Step 1: \blue{Automated agent verification}} As a reviewer, SAR-Agent interacts with the database to verify the annotation and generates a diagnostic report. If the report confirms that the annotation is correct, it is accepted. Otherwise, SAR-Agent returns the report to the annotator, including any detected issues.
    
\minihead{Step 2: \blue{Manual review and adjudication}} The annotators review the reported issues. If they agree with the issues, they revise the annotation. If they disagree, they consult a SQL expert reviewer for a final decision. If the expert considers the annotation correct, it is accepted; otherwise, the annotators are required to revise it.

\minihead{Step 3: Annotation revision} \blue{When SAR-Agent detects errors, human annotators revise the original annotations using the diagnostic report, which details error reasons and proposes a candidate query revision. For the SQL query, they may adopt the revision suggested by SAR-Agent after manual verification. If they disagree with the proposal, they need to revise the query themselves. For the user question and external knowledge, annotators need to manually correct any errors to ensure accuracy and natural, idiomatic phrasing. After revision, the workflow returns to Step 1.}


\blue{We used SAPAR to correct annotation errors in existing text-to-SQL benchmarks. When researchers develop a new text-to-SQL benchmark, they can apply SAPAR to improve annotation quality after creating the initial annotations.}
\section{BIRD and Spider 2.0 Have Pervasive Annotation Errors}
\label{sec:error_dectection}
We conducted a comprehensive audit of two widely studied text-to-SQL benchmarks: BIRD Mini-Dev \cite{bird} and Spider 2.0-Snow \cite{spider2}. In this section, we first describe our three-step examination procedure. We then report the error rate for each benchmark. \blue{Finally, we analyze all detected annotation errors based on four patterns introduced in Section \ref{sec:introduction}.}

\subsection{Experimental Settings} 
\minihead{Data settings}
We evaluated two widely-studied text-to-SQL benchmarks: BIRD Mini-Dev ~\cite{bird} and Spider 2.0-Snow~\cite{spider2}.

\begin{enumerate}[leftmargin=*]
\item \textit{BIRD Mini-Dev:} Following prior work \cite{BIRDminidev}, we use the first version of BIRD Mini-Dev, which contains 500 \texttt{SELECT}-only text-to-SQL examples carefully chosen by the BIRD team from the original BIRD Dev set.\footnote{The BIRD Mini-Dev dataset is available at \url{https://github.com/bird-bench/mini_dev/tree/main/llm/mini_dev_data}.} However, prior work reports two duplicate examples, leaving 498 unique examples \cite{BIRDminidev}. 
We find that the databases provided in the BIRD Mini-Dev set contain fewer rows compared to those in the BIRD Dev set. To better evaluate the correctness of SQL queries, we use databases from the BIRD Dev set in our experiments. 
\item \textit{Spider 2.0-Snow:} Spider 2.0-Snow is a variant of Spider 2.0. We choose this variant because teams at leading technology companies benchmark their agents on it rather than on Spider 2.0-lite. Spider 2.0-Snow contains a total of 547 examples, with gold queries publicly available for 121 of them. Therefore, we evaluated the correctness of these 121 examples.

\end{enumerate}

\minihead{Metrics}
We define the error rate of each text-to-SQL benchmark as the percentage of incorrect examples in the benchmark.
We consider an example to be incorrect if it exhibits any of \textbf{E1} - \textbf{E4}.

\minihead{Agent settings}
We ran SAR-Agent with the OpenAI o3 model \cite{o3} and configured the agent to complete the annotation error diagnostic within a maximum of 30 iterations.\footnote{We release SAR-Agent's prompt and scaffolding at \url{https://github.com/uiuc-kang-lab/text_to_sql_benchmarks/tree/main/SAR-Agent}.}

\subsection{Three-step Examination}
\label{subsec:exam}
Our three-step examination procedure consists of automatic detection, manual verification, and additional manual detection. We detail each step below.

    \minihead{Automatic detection} We first executed \agent on two text-to-SQL benchmarks: BIRD Mini-Dev \cite{bird} and Spider 2.0-Snow \cite{spider2}. For each example, \agent generated a diagnostic report.
    
    \minihead{Manual verification} 
    After generating the diagnostic report, we manually verified the correctness of error reasons for all examples that \agent labels as incorrect or ambiguous. Two authors independently assessed each example to ensure the accuracy of the verification process. In instances of disagreement, all four authors collectively reviewed and discussed the examples to reach a consensus on whether an example should be classified as incorrect. Our manual verification process consists of two main parts. First, for questions requiring domain-specific knowledge, we consulted authoritative online resources to cross-check the explanations produced by the LLM against the definitions provided by human annotators. For example, in question 46, the annotator asserted that ``grades 1-12 means K-12.'' However, \agent identified this interpretation as incorrect, noting that ``K-12'' includes kindergarten. To validate this assessment, we referred to the definition of K-12 on Wikipedia,\footnote{\url{https://en.wikipedia.org/wiki/K-12}} which aligns with the judgment of SAR-Agent. Second, for each annotated query, we executed a set of verification queries, such as subqueries or modified queries based on the listed error reasons, on the relevant database to assess the correctness of the SAR-Agent's judgment.

     \minihead{Additional manual detection} Through manual verification, we identified recurring error patterns across different examples. To further investigate annotation errors, we examined other examples within the same database to determine whether similar errors are present elsewhere \cite{osti_10316595}. For BIRD Mini-Dev, we also combined the errors identified in prior work \cite{BIRDminidev}.


\begin{table}[t]
\caption{\blue{Error pattern distribution in BIRD Mini-Dev and Spider 2.0-Snow. Each example may contain multiple errors across different patterns.}}
\label{tab:stat}
    \centering
    \small
    \begin{tabular}{ccc}
        \toprule
        \multirow{2}{*}{\textbf{Error pattern}}  & \multicolumn{2}{c}{\textbf{\# Examples}} \\
        \cline{2-3}
        
         & \textbf{BIRD Mini-Dev} & \textbf{Spider 2.0-Snow} \\

        \midrule
        \textbf{E1} & 77 (29.28\%) & 37 (48.68\%) \\
        \textbf{E2} & 152 (57.79\%) & 44 (57.89\%) \\
        \textbf{E3} & 28 (10.65\%) & 9 (11.84\%) \\
        \textbf{E4} & 78 (29.66\%) & 18 (23.68\%) \\
        \midrule
         \makecell[l]{\textbf{Examples with}\\\textbf{annotation errors}} & 263  & 76 \\
        \bottomrule

    \end{tabular}
\end{table}

\begin{table}[t]
\caption{Representative examples for each error pattern in BIRD Mini-Dev. The question IDs in brackets indicate questions that were marked as noisy in prior work, but due to different types of errors.}
    \label{tab:bird}
    \small
    \centering
    \renewcommand{\arraystretch}{1.2} 
    \begin{tabular}{|l|c|p{4.1 cm}|}
        \hline
        \textbf{\makecell*[l]{Question IDs}} & \textbf{\makecell*[l]{Error\\Pattern}} & \textbf{Explanation}  \\ \hline
        \small\texttt{92}, \small\texttt{149} & E1 & \texttt{BETWEEN ... AND ...} includes the boundary values, it conflicts with the question's requirement for exclusive comparisons. \\ \hline 
         \makecell*[l]{\small\texttt{12}, \small\texttt{36}, \small\texttt{40}, \small\texttt{50},\\ (\small\texttt{17})}  & E2 & \makecell*[l]{The query does not include the\\ restriction \texttt{rtype = `S'}.} \\ \hline 
         \makecell*[l]{\small\texttt{633}, \small\texttt{639}, \small\texttt{640},\\ \small\texttt{685}, (\small\texttt{634}, \small\texttt{637})}  & E2 & \makecell*[l]{The queries consider users who ever\\ edited rather than the user who\\ owns the post.} \\ \hline 
         \small\texttt{1376}, \small\texttt{1378}, \small\texttt{1403}  & E2 & \makecell*[l]{No aggregation over events.} \\ \hline 
         \small\texttt{32}, \small\texttt{46}, \small\texttt{62}  & E3 & \makecell*[l]{``K-12'' is not equivalent to grades 1 \\ through 12.} \\ \hline 
         \makecell*[l]{\small\texttt{862}, \small\texttt{877}, \small\texttt{881},\\ \small\texttt{940}, \small\texttt{954}} & E3 & \makecell*[l]{Drivers with a ``+n Lap'' status also \\ finished the race.} \\ \hline 
         \makecell*[l]{ \small\texttt{1156},  \small\texttt{1229}, \small\texttt{1235},\\(\small\texttt{1168}, \small\texttt{1171}, \small\texttt{1227},\\ \small\texttt{1232}, \small\texttt{1239}, \small\texttt{1243})} & E3 & \makecell*[l]{The age calculation is not  accurate \\ for  patients.} \\ \hline 
         \small\texttt{1175} & E4 &  \makecell*[l]{``Doctor’s diagnosis'' could come \\ from either Examination.Diagnosis \\ or Patient.Diagnosis.} \\ \hline 
         
    \end{tabular}
\end{table}


\subsection{Evaluation Results}
Using the three-step examination, we identify annotation errors in 263 of the 498 (52.8\%) BIRD Mini-Dev examples, including 161 examples reported in prior work \cite{BIRDminidev} and additional errors uncovered by our pipeline. For Spider 2.0-Snow, we find annotation errors in 76 of 121 examples, resulting in an error rate of 62.8\%.

\subsection{Analysis of Annotation Errors}
\label{subsec:anno_issue}
\blue{We begin by detailing the classification criteria for each error pattern, then present their distributions in BIRD Mini-Dev and Spider 2.0-Snow, and conclude with representative examples drawn from both benchmarks.
}

\minihead{Classification criteria for error patterns}
\blue{We detail the criteria used to classify the four error patterns introduced in Section~\ref{sec:introduction}. Given a natural language input $\mathcal{T}$, a ground-truth SQL query $\mathcal{Q}$, and the corresponding database $\mathcal{D}$:}

\begin{enumerate}[label=\textbf{E\arabic*.},leftmargin=*]
\item \blue{We flag \textbf{E1} by comparing $\mathcal{T}$ and $\mathcal{Q}$ without accessing $\mathcal{D}$. Typical cases include missing or extraneous filter predicates in $\mathcal{Q}$, incorrect comparison operators (e.g., $\mathcal{T}$ specifies “strictly greater than 100” while $\mathcal{Q}$ uses “$\geq 100$”), or an incorrect formula (e.g., $\mathcal{T}$ asks for “\% faster” but $\mathcal{Q}$ computes $(\texttt{speed853} - \texttt{speed854})/\texttt{speed853} \times 100$).}

\item  \blue{Detecting \textbf{E2} requires reasoning over $\mathcal{D}$ (both schema and data). We examine the schema and data and execute queries to flag \textbf{E2}. Examples include joining on a non-unique key without the necessary deduplication, missing necessary aggregation, or ignoring required format conversions.}

\item \blue{We flag \textbf{E3} when the annotation relies on domain knowledge that is incorrect or inconsistent with authoritative sources. We verify such knowledge by consulting credible documentation and cross-checking references.}

\item \blue{We flag \textbf{E4} when the user question in $\mathcal{T}$ admits multiple reasonable interpretations that map to different, non-equivalent SQL queries producing different results over $\mathcal{D}$, or when the desired output specification is underspecified.}
\end{enumerate}

\minihead{Distribution of error patterns} \blue{Following our classification criteria, we computed the distribution of error patterns in BIRD Mini-Dev and Spider 2.0-Snow. Because an example can contain multiple errors, it may be assigned to multiple patterns. As shown in Table \ref{tab:stat}, \textbf{E2}, attributable to a limited understanding of the data or the schema, is the most common error pattern in both benchmarks. Among examples containing annotation errors, \textbf{E2} occurs in 57.79\% of BIRD Mini-Dev examples and 57.89\% of Spider 2.0-Snow examples.
}

\begin{table}[t]
\caption{Representative examples for each error pattern in Spider 2.0-Snow.}
    \label{tab:snow}
    \small
    \centering
    \renewcommand{\arraystretch}{1.2} 
    \begin{tabular}{|l|c|p{4cm}|}
        \hline
        \textbf{Question IDs} & \textbf{\makecell*[l]{Error\\Pattern}} & \textbf{Explanation}  \\ \hline 
        
        \makecell*[l]{\small\texttt{sf\_bq263}, \small\texttt{sf\_bq271},\\\small\texttt{sf\_bq273}, \small\texttt{sf\_bq294}}  & E1 & \makecell*[l]{The function TO\_TIMESTAMP\\(end\_date) sets the time to the \\ start  of the end date (00:00:00.000),\\ instead of to the end of the day \\(23:59:59.999) as required.} \\ \hline 
        \makecell*[l]{ \small\texttt{sf\_bq050},\small\texttt{sf\_bq209}, \\\small\texttt{sf\_bq455}}  & E1 & \makecell*[l]{The SQL queries lack the required \\ filters specified by the question.} \\ \hline 
        \makecell*[l]{\small\texttt{sf\_bq099}, \small\texttt{sf\_bq248},\\ \small\texttt{sf\_bq422},\\ \small\texttt{sf\_local263}}  & E2 & \makecell*[l]{Using \texttt{JOIN} or \texttt{FLATTEN} operators \\ inflates row counts; the query \\ lacks  \texttt{DISTINCT} for deduplication.} \\ \hline 
        \small\texttt{sf\_bq052}, \small\texttt{sf\_bq246}  & E3 & \makecell*[l]{Forward citations are miscalculated.} \\ \hline 
        \makecell*[l]{\small\texttt{sf\_bq017}, \small\texttt{sf\_bq068},\\
        \small\texttt{sf\_bq099},
        \small\texttt{sf\_bq182},\\ 
        \small\texttt{sf\_bq193},
        \small\texttt{sf\_bq222},\\ 
        \small\texttt{sf\_bq223}}  & E4 & \makecell*[l]{The gold SQL queries apply \\ additional  quotation mark \\  replacements that the question \\  does  not specify.} \\ \hline 
    \end{tabular}
\end{table}

\minihead{Representative Errors in BIRD Mini-Dev}
\label{subsec:bird}
We provide representative, newly discovered examples from BIRD Mini-Dev for each of the four error patterns in Table~\ref{tab:bird} and discuss them in more detail.
\begin{enumerate}[label=\textit{Examples of \textbf{E\arabic*.}}, wide=0pt]
\item The annotators used the inclusive operator BETWEEN ... AND ... in the gold query, but the question specifies strict inequalities (>, <) that exclude the boundary values.  Consequently, the logic of these queries is incorrect.

\vspace{0.5em}

\item  We find three typical errors for \textbf{E2}. Firstly, in the five listed questions related to the \texttt{california\_schools} database, we find the annotated queries omit the predicate \texttt{rtype = `S'}. Although annotators labeled the \texttt{rtype} column as ``unuseful'' in the table schema, our investigation on the database confirms that \texttt{rtype} is useful as it distinguishes schools from districts. Therefore, for questions referencing schools, the ground truth SQL queries should include \texttt{rtype = `S'}.

\setlength{\parindent}{9pt}

\indent Secondly, in the \texttt{codebase\_community} database, the \texttt{posts} table records the owner of each post, while the \texttt{postHistory} table tracks the editors. In the six listed questions, phrases such as ``posted by'' or ``his post'' should refer to the post owner, not to the editors.

\indent Thirdly, in the \texttt{student\_club} database, each event can have multiple corresponding rows in the \texttt{budget} table. Therefore, when calculating the budget for an event, the query should aggregate the budget values associated with that event.

\vspace{0.5em}

\item We find three typical errors for \textbf{E3}. Firstly, in the three listed examples in Table ~\ref{tab:bird} from the \texttt{california\_schools} database, the annotators mistakenly interpreted ``K-12'' as referring solely to ``grades 1–12,'' thereby excluding kindergarten.

\indent Secondly, in the context of Formula 1, ``+n Lap'' indicates that the driver finished the race but was n laps behind the race leader. Due to a lack of domain knowledge, the annotator ignored these drivers when writing the gold query.

\indent Thirdly, according to the Centers for Disease Control and Prevention (CDC), a patient's age should be calculated as the number of complete years that have passed since the patient's date of birth \cite{age}. The formula provided in the evidence simply subtracts the year of birth from the current year, which is not accurate.

\vspace{0.5em}

\item In the \texttt{thrombosis\_prediction} database, both the \texttt{Examination} and \texttt{Patient} tables contain the \texttt{Diagnosis} column. However, question 1175 asks for the ``Doctor's diagnosis,'' and neither the question nor the provided external knowledge specifies which table should be referenced.

\end{enumerate}

\minihead{Representative Errors in Spider 2.0-Snow}
We identify annotation errors in 76 out of the 121 examples in Spider 2.0-Snow that have released gold queries. We show representative error examples for each error pattern in Table \ref{tab:snow}, and discuss the details below:
\begin{enumerate}[label=\textit{Examples of \textbf{E\arabic*.}}, wide=0pt]
\item We identify two typical errors for the error pattern \textbf{E1}. Firstly, in four examples, we find that Spider 2.0 annotators used \texttt{BETWEEN TO\_TIMESTAMP(start\_date) AND TO\_TIMESTAMP(end\_\\date)}. 
However,  \texttt{TO\_TIMESTAMP()} casts a date to the timestamp of the start of the day (00:00:00.000) rather than the end (23:59:59.999). Accordingly, the gold queries omit rows on the end date that occur after midnight, resulting in the wrong query results.

\setlength{\parindent}{9pt}

\indent Secondly, we find that three annotated queries omit the required restrictions specified in the question. For example, in \textit{sf\_bq209}, the question asks for a ``utility patent,'' but the annotated query does not include this restriction.

\vspace{0.5em}

\item The annotators did not filter or deduplicate the intermediate results after the \texttt{JOIN} or \texttt{FLATTEN} operations, which resulted in row inflation in four examples. 

\vspace{0.5em}

\item We find that  two annotated SQL queries involving forward citation calculation use incorrect join conditions. The queries mistakenly join on \texttt{cited.patent\_id = apps.patent\_id} instead of \texttt{cited.citation\_id = apps.patent\_id}.\footnote{In sf\_bq128, the annotators used the correct join key.}

\vspace{0.5em}

\item We identify seven questions with ambiguous output format specifications. Agents must generate SQL queries that replace quotation marks with the empty string (`') to pass the evaluation. Otherwise, the evaluator marks all answers incorrect. However, the questions do not state this requirement.
\end{enumerate}








\begin{figure}[t]
  \begin{subfigure}[t]{\columnwidth}
   \input{figures/bird985}
\caption{Example 985 of BIRD Dev. The query lacks data type conversion (\textbf{E2}), and the output column is ambiguous (\textbf{E4}).}
\label{sub:bird_issue_2}
\end{subfigure}

 \vspace{1em}

  \begin{subfigure}[t]{\columnwidth}
    \input{figures/bird985_fix}
\caption{Corrected example 985 in BIRD Dev by correcting the user question and the ground-truth SQL query. We also update the database to prevent incorrect queries from producing the same results.}
\label{sub:spider2_issue_2}
\end{subfigure}

\caption{Original and corrected example 985 in BIRD Dev.}
\label{fig:fixed_sql}
\end{figure}

\section{Annotation Errors Break the BIRD Leaderboard}
\label{sec:reevaluate}
The leaderboards of text-to-SQL benchmarks provide guidance for both text-to-SQL research and practical deployment. Researchers and practitioners use agent rankings to understand technique gaps and choose agents for downstream applications. To understand the impact of annotation errors in text-to-SQL benchmarks on agent performance and rankings, we evaluated all open-source text-to-SQL agents on the BIRD leaderboard whose reported results are reproducible in local environments. Our evaluation demonstrates the unreliability of the BIRD leaderboard.


In this section, we first describe how we corrected a subset of BIRD and then detail the experimental settings. We next present changes in execution accuracy and rankings of all evaluated agents. We analyze the causes of these performance variations to understand how annotation errors affect text-to-SQL agents.
Finally, we analyze two rank correlations: (i) between the agent rankings on the original Dev subset and the full Dev set, and (ii) between the agent rankings on the original and corrected Dev subsets.

\subsection{Data Settings}
\minihead{Dataset} We randomly sampled 100 examples from the BIRD Dev set. 
As shown in Table~\ref{tab:fixed_stat}, our selection includes 62 simple, 28 moderate, and 10 challenging examples.

\minihead{Benchmark Correction}
We manually corrected the sampled Dev subset based on SAPAR. For each text-to-SQL example, according to the diagnostic report generated by SAR-Agent, we made the following corrections.
First, following prior work \cite{BIRDminidev}, we corrected all identified issues by revising natural language questions, external knowledge, and ground truth SQL queries to generate a corrected Dev subset.
Second, as mentioned in Section~\ref{subsec:bird}, the value description of the \texttt{rtype} column was previously marked as ``unuseful.'' We corrected the schema to fix this issue in the revised Dev subset.
Finally, we found examples where incorrect queries produced the same results as the corrected queries due to limitations in the underlying data. To enable more accurate evaluation, we addressed this issue by inserting distinguishing records in the database. We conducted up to three iterations of review and revision to reach consensus with SAR-Agent for each text-to-SQL pair.

To understand our correction process, we use the Example 985 of BIRD Dev as an example (Figure \ref{fig:fixed_sql}). We found that both the original user query (in natural language) and gold SQL query are incorrect. Therefore, we corrected both the user query and SQL query. Furthermore, we identified that the original data failed to expose errors of incorrectly ranking time. To address this, we inserted a row with the \texttt{time} column set to ``11:00.365'' into the database.

\minihead{Statistics of Correction} As shown in Table~\ref{tab:fixed_stat}, we corrected 48\% of the examples. Specifically, we corrected the questions in 19 examples, addressed external knowledge issues in 17 examples, corrected the SQL queries in 41 examples, and fixed the schema in one example. In addition, we also modified the data in the database for six examples to prevent incorrect SQL logic from being mistakenly considered correct. 

\begin{table}[t]
\caption{We corrected a total of 48 text-to-SQL examples. Each example may involve more than one type of correction.}
\label{tab:fixed_stat}
    \centering
    \small
    \begin{tabular}{lc}
        \toprule
        \textbf{Statistics} & \makecell[l]{\textbf{\# Examples} \\ {(total=100)}}\\ 
        \midrule
        \textbf{Difficulty of selected examples} &  \\
        \hspace{1em} Easy & 62 \\ 
        \hspace{1em} Moderate & 28 \\ 
        \hspace{1em} Challenging & 10 \\
        \midrule
        \textbf{Correction type of Dev subset } &  \\ 
        Corrected examples  & 48  \\
        \hspace{1em} Question correction& 19  \\ 
        \hspace{1em} External knowledge correction & 17 \\ 
        \hspace{1em} SQL correction & 41 \\ 
        \hspace{1em} Schema correction & 1 \\ 
        \hspace{1em} Database modification& 6 \\ 
        \bottomrule
    \end{tabular}
\end{table}

\subsection{Experimental Settings}


\minihead{Metrics}
Following prior work~\cite{bird}, we use Execution Accuracy (EX) and ranking as our metrics. EX determines correctness by comparing execution results between predicted SQL queries and ground truth queries on the database. The ranking metric determines each text-to-SQL agent's position among all candidates. We further use Spearman's rank correlation coefficient ($r_s$)~\cite{spearman} for rank correlation analysis, which measures monotonic relationships between rank variables, ranging from $-1$ to $+1$. 

\minihead{Text-to-SQL Agents}
\blue{We evaluated all 16 agents (Contextual-SQL~\cite{agrawal2025text2sql}, CSC-SQL~\cite{cscsql}, GenaSQL~\cite{genasql}, OpenSearch-SQL~\cite{opensearchsql}, OmniSQL~\cite{omnisql}, CHESS~\cite{chess}, Alpha-SQL~\cite{li2025alphasqlzeroshottexttosqlusing}, GSR~\cite{gsrsql}, RSL-SQL~\cite{cao2024rsl}, E-SQL~\cite{esql}, TA-SQL~\cite{qu-etal-2024-generation}, DTS-SQL~\cite{pourreza2024dtssqldecomposedtexttosqlsmall}, MAC-SQL~\cite{wang2025macsqlmultiagentcollaborativeframework}, SFT CodeS-15B~\cite{codes}, DAIL-SQL~\cite{gao2023texttosqlempoweredlargelanguage}, DIN-SQL~\cite{pourreza2023dinsqldecomposedincontextlearning}) on the BIRD leaderboard that have publicly available execution code and can be reproduced in local environments.\footnote{We chose agents based on leaderboard rankings as of August 20, 2025.} We show the detailed configurations of these agents in Appendix \ref{subsec:config_agent}.}



\begin{figure*}[t]
    \centering
    \includegraphics[width=1\linewidth]{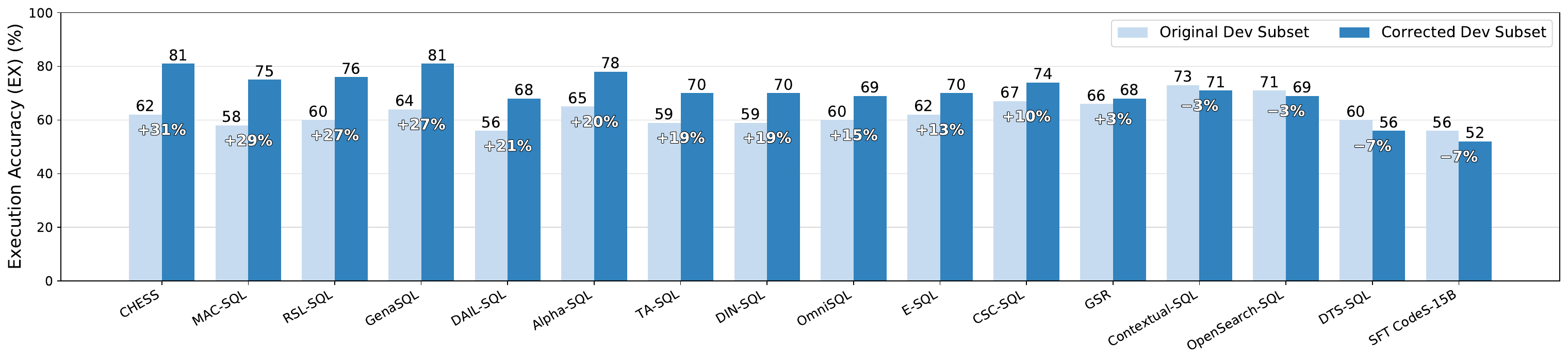}
    \caption{Execution accuracy of agents on original and corrected BIRD Dev subsets. The agents' performance changes range from -7\% to 31\% in relative terms.
    }
    \label{fig:ex}
\end{figure*}

\begin{figure*}[t]
    \centering
    \includegraphics[width=1\linewidth]{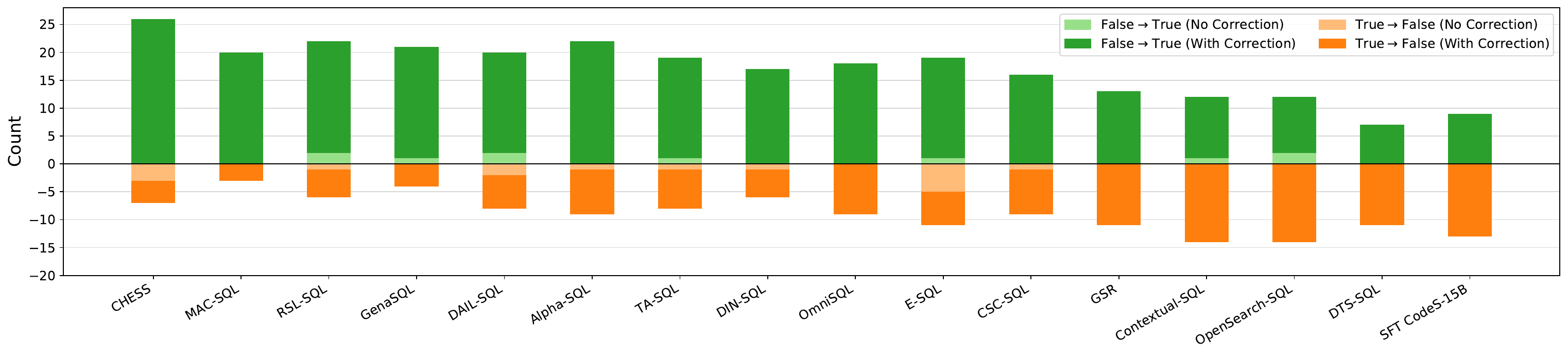}
    \caption{Number of examples with correctness changes between original and corrected BIRD Dev subsets.}
    \label{fig:ex_cnt}
\end{figure*}

\subsection{Evaluation Results}
We executed all 16 agents on both the original and corrected BIRD Dev subsets to evaluate how annotation errors affect agent performance. We will discuss the change of execution accuracy and rankings of all evaluated agents in the following paragraphs. 

\minihead{Execution Accuracy (EX)}
As shown in Figure~\ref{fig:ex}, the relative changes in EX across the agents range from $-7$\% to $31$\%. Specifically, 12 out of 16 agents exhibit improved EX after correction. Among these, CHESS demonstrates the most significant increase, with EX rising from 62\% to 81\%, a 30.6\% relative improvement. In contrast, four agents experience a performance decline. SFT CodeS-15B shows the most significant decrease, with EX dropping from 56\% to 52\%, corresponding to a 7.1\% relative decline. We provide a detailed analysis for the performance change of CHESS and SFT CodeS-15B in Section \ref{subsec:perf-change-analysis}.

We then analyze the  factors responsible for the changes in EX. We present the number of examples for which correctness changed, alongside information on whether annotation corrections were applied in Figure~\ref{fig:ex_cnt}. Across all 16 agents, annotation corrections change their correctness on an average of 16.4 examples from \texttt{False} to \texttt{True} and 8.0 from \texttt{True} to \texttt{False}. Among the 16 agents, CHESS shows the largest number of \texttt{False} to \texttt{True} updates (26), while DTS-SQL shows the fewest (7). In contrast, for \texttt{True} to \texttt{False} changes, Contextual-SQL and OpenSearch-SQL tie for the most (14), and MAC-SQL has the fewest (3).

\begin{figure}[t]
    \centering
    \includegraphics[width=\linewidth]{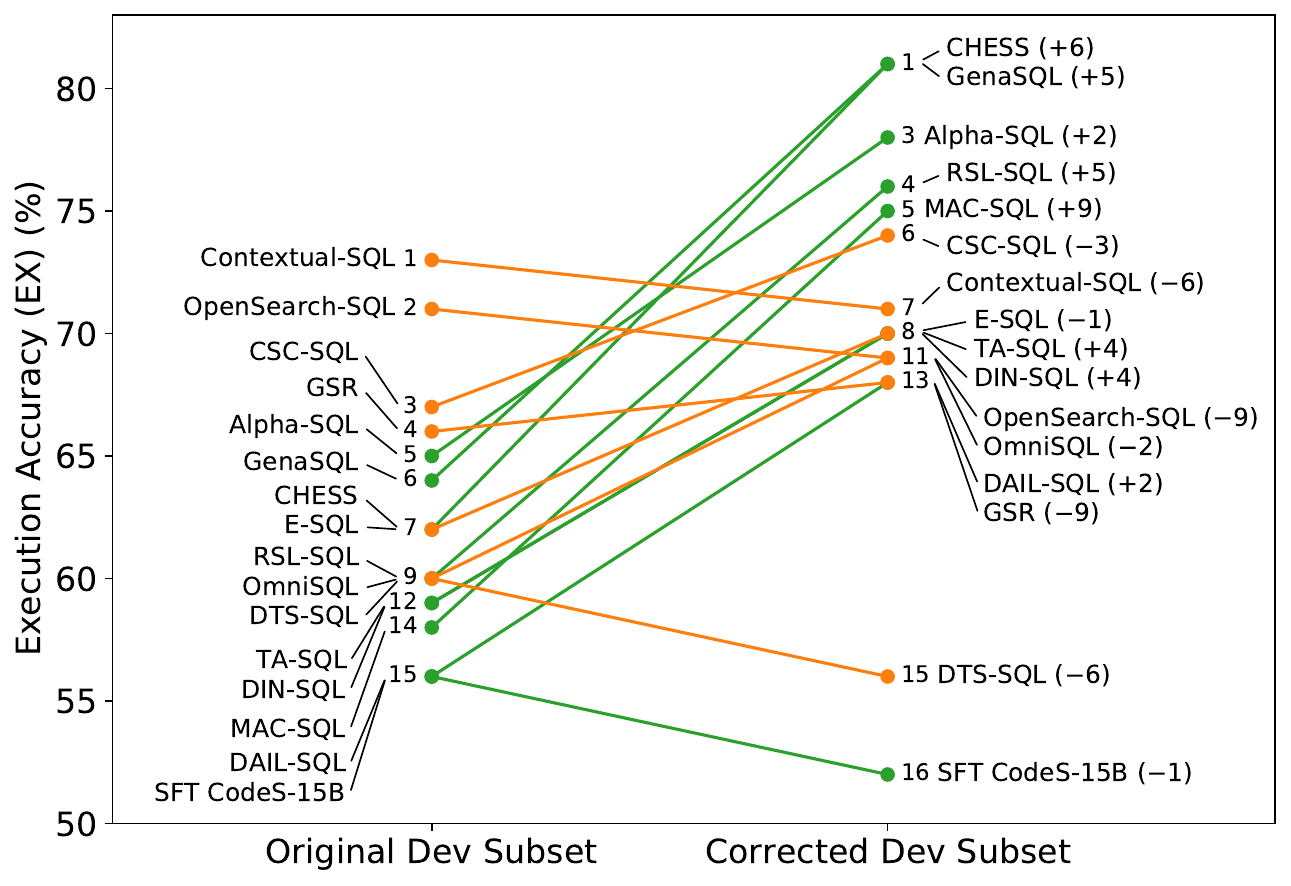}
    \caption{Agent ranking changes from original to corrected BIRD Dev subsets. Ranking shifts range from -9 to +9 positions, with an average change of 5 positions.}
    \label{fig:ranking}
\end{figure}

\minihead{Rankings}
We compare agent rankings before and after correction and find significant changes in their relative positions. As shown in Figure~\ref{fig:ranking}, CHESS and GenaSQL, which achieve substantial relative performance improvements of 30.6\% and 26.6\% respectively, elevate from middle-ranked positions (7th and 6th place) to tie for first place with 81\% EX. MAC-SQL shows the most dramatic ranking shift, moving 9 positions from 14th to 5th place due to the second-highest relative performance gain of 29.3\%. In contrast, GSR experiences the sharpest decline, falling 9 positions from 4th to 13th place. Despite achieving a 3\% relative improvement in performance, GSR's marginal gain becomes insignificant compared to other improving agents, which average 21\% relative improvement.

All agents exhibiting a decline in EX experience ranking drops. Previous top-ranking agents, Contextual-SQL and OpenSearch-SQL, suffer notable repositioning due to performance degradation. Contextual-SQL falls 6 positions from 1st to 7th place, while OpenSearch-SQL drops 9 positions from 2nd to 11th place. In addition, DTS-SQL and SFT CodeS-15B also decline by 6 and 1 positions, respectively. 

Across all 16 agents, ranking changes average 5 positions, with individual shifts ranging from $-9$ to $+9$ positions. These results demonstrate that rankings of methods are highly sensitive to the annotation errors in the benchmark. High error rates present in the benchmark can significantly undermine the reliability of a leaderboard, which fails to accurately represent the relative performance of different agents. Consider a practitioner seeking to select the best text-to-SQL agent for a downstream application based on the BIRD leaderboard. The flawed leaderboard can mislead this decision and lead to the selection of a suboptimal agent.

\subsection{Analysis of Performance Changes} \label{subsec:perf-change-analysis}

In this subsection, we analyze why agents experience varying degrees of performance change from annotation corrections by investigating CHESS \cite{chess} and SFT CodeS-15B \cite{codes}, the agents showing the largest improvement and the largest degradation.

\miniheadnodot{Why does the EX of CHESS increase?} 
As shown in Table~\ref{tab:chess}, we identify 12 examples where correcting errors only in $\mathcal{Q}$ (the gold query) changes the evaluation status from \texttt{False} to \texttt{True} for CHESS. We analyze the CHESS-generated queries for these 12 examples in the original Dev set. We find that 91.7\% of generations match the revised ground truth, indicating that the original \texttt{False} labels are false negatives. Across these 12 examples, we find a common error pattern: in 5/12 (41.7\%) of the examples (\texttt{310}, \texttt{416}, \texttt{605}, \texttt{1286}, \texttt{1302}), human annotators omitted the required \texttt{DISTINCT} keyword within the \texttt{COUNT} function, leading to incorrect \texttt{False} classifications in the original evaluation. 
In addition, among the examples that require revising both $\mathcal{T}$ (the question or external knowledge) and $\mathcal{Q}$ (the gold query) to resolve ambiguities and semantic inconsistencies, we find that 14 CHESS-generated queries are reclassified as \texttt{True}.

\begin{table}[t]
\caption{Examples with correctness changes for CHESS categorized by annotation correction type.}
    \label{tab:chess}
    \small
    \centering
    \renewcommand{\arraystretch}{1.2}
    
    \begin{tabular}{|l|c|p{4.5cm}|}
        \hline
        \textbf{Category} & & \textbf{Question IDs} \\ \hline
         Corrected $\mathcal{T}$ & T $\rightarrow$ F & \small\texttt{255} \\ \hline
         \multirow{2}{*}{Corrected $\mathcal{Q}$} & F $\rightarrow$ T & \small\texttt{310}, \small\texttt{416},  \small\texttt{442}, \small\texttt{484}, \small\texttt{605}, \small\texttt{610}, \small\texttt{646}, \small\texttt{888}, \small\texttt{928}, \small\texttt{1200}, \small\texttt{1286}, \small\texttt{1302} \\ \cline{2-3}
         & T $\rightarrow$ F & \small\texttt{42} \\ \hline
        \multirow{2}{*}{Corrected $\mathcal{T}$ \& $\mathcal{Q}$} & F $\rightarrow$ T & \small\texttt{180}, \small\texttt{305},
        \small\texttt{406},
        \small\texttt{428},
        \small\texttt{602}, \small\texttt{620}, \small\texttt{648}, \small\texttt{772}, \small\texttt{855}, \small\texttt{970}, \small\texttt{987}, \small\texttt{1004}, \small\texttt{1271}, \small\texttt{1280} \\ \cline{2-3}
         & T $\rightarrow$ F & \small\texttt{829}, \small\texttt{1173} \\ \hline
    \end{tabular}
\end{table}

\begin{table}[t]
\caption{Examples with correctness changes for SFT CodeS-15B categorized by annotation correction type.}
    \label{tab:codes}
    \small
    \centering
    \renewcommand{\arraystretch}{1.2}
    
    \begin{tabular}{|l|c|p{4cm}|}
        \hline
        \textbf{Category} & & \textbf{Question IDs} \\ \hline
         \multirow{2}{*}{Corrected $\mathcal{Q}$} & F $\rightarrow$ T & \small\texttt{442}, \small\texttt{605}, \small\texttt{610}, \small\texttt{646}, \small\texttt{888}, \small\texttt{928} \\ \cline{2-3}
         & T $\rightarrow$ F & \small\texttt{1200}, \small\texttt{1286}, \small\texttt{1302} \\ \hline
        \multirow{2}{*}{Corrected $\mathcal{T}$ \& $\mathcal{Q}$} & F $\rightarrow$ T & \small\texttt{406}, \small\texttt{855}, \small\texttt{1271} \\ \cline{2-3}
         & T $\rightarrow$ F & \small\texttt{305}, \small\texttt{620}, \small\texttt{648}, \small\texttt{772}, \small\texttt{829}, \small\texttt{846}, \small\texttt{935}, \small\texttt{987}, \small\texttt{1173}, \small\texttt{1280} \\ \hline
    \end{tabular}
\end{table}


\miniheadnodot{Why does the EX of SFT CodeS-15B decrease?}
As shown in Table~\ref{tab:codes}, after correcting $\mathcal{Q}$ (the gold query), six examples flip from \texttt{False} to \texttt{True}, whereas three flip from \texttt{True} to \texttt{False}. Of the latter three, two (\texttt{1286}, \texttt{1302}) change due to the agent committed the same error as the human annotator (missing \texttt{DISTINCT} keyword issue in \texttt{COUNT} function). 
For examples requiring revisions to both $\mathcal{T}$ (the question or external knowledge) and $\mathcal{Q}$ (the gold query), only three switch from \texttt{False} to \texttt{True}, whereas ten switch from \texttt{True} to \texttt{False}.

\begin{figure}[t]
    \centering
    \begin{subfigure}[t]{\linewidth}
        \centering
        \includegraphics[width=0.8\linewidth]{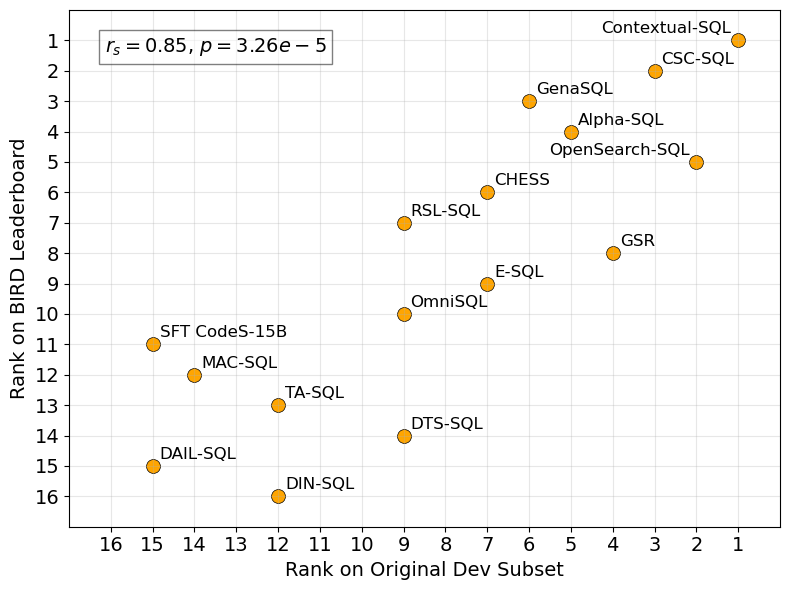}
        \caption{Correlation between EX rankings on the original Dev subset and the full BIRD Dev set. $r_s = 0.85$ indicates a strong positive correlation.}
        \label{fig:rank_corr_1}
    \end{subfigure}%
    \vspace{1em}
    \begin{subfigure}[t]{\linewidth}
        \centering
        \includegraphics[width=0.8\linewidth]{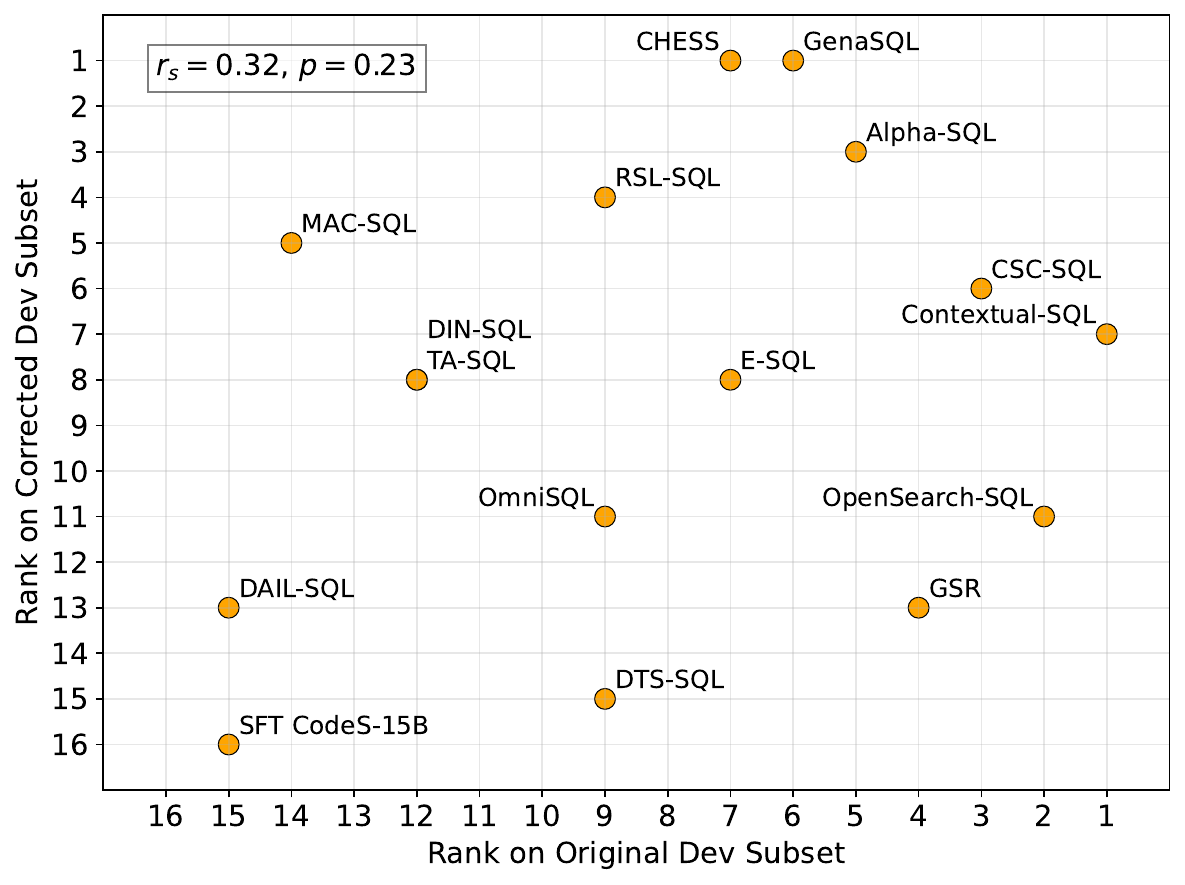}
        \caption{Correlation between EX rankings on the original and corrected Dev subset. $r_s = 0.32$ indicates weak correlation.}
        \label{fig:rank_corr_2}
    \end{subfigure}
    \caption{Ranking Correlation Analysis using Spearman's Rank Correlation Coefficient $r_s$.}
    \label{fig:rank_corr}
\end{figure}

\subsection{Analysis of Ranking Correlations}
To assess the impact of annotation errors on rankings, we conducted two correlation analyses using Spearman's rank correlation coefficient~\cite{spearman}.
We first examined the correlation between EX rankings on the original Dev subset and the BIRD leaderboard (Dev). As shown in Figure~\ref{fig:rank_corr_1}, the coefficient $r_s$ is 0.85 with $p$-value less than 0.001. This strong positive correlation confirms that the relative performance characteristics of agents on the full BIRD Dev set are preserved in our randomly sampled examples.

We also compared rankings on the original Dev subset and our corrected Dev subset, where the only factor that differs is whether annotation errors are corrected. As shown in Figure~\ref{fig:rank_corr_2}, the coefficient $r_s$ is 0.32 with a $p$-value of 0.23. This weak correlation indicates that the presence of annotation errors substantially affects ranking outcomes, compromising the reliability of rankings of agents on the original dataset.


\section{SAR-Agent helps experts detect 42\% more incorrect examples}
\label{sec:SAR}
We evaluated the ability of SAR-Agent in detecting text-to-SQL annotation errors. In this section, we first describe the precision of SAR-Agent on BIRD Mini-Dev and Spider 2.0-Snow. We then compare the number of detected error examples by SAR-Agent with those by the Arcwise team. Finally, we present the per-example monetary cost and the number of execution steps of SAR-Agent.

Building upon the corrections by Arcwise, we addressed additional errors identified in BIRD Mini-Dev \cite{BIRDminidev}. To facilitate fair evaluation, we have released two versions of the dataset: (1) Arcwise-Plat-SQL: we only corrected the SQL annotations, preserving original ambiguities in the questions and evidence; (2) Arcwise-Plat: we resolved both incorrect SQL annotations and underlying ambiguities. Both versions are available in our GitHub repository.\footnote{\url{https://github.com/uiuc-kang-lab/text_to_sql_benchmarks/tree/main/data}}

\subsection{Experimental Settings}
\minihead{Data settings} We analyze SAR-Agent’s performance on BIRD Mini-Dev and Spider 2.0-Snow based on the generated diagnostic reports in error detection experiments (Section ~\ref{sec:error_dectection}).

\minihead{Metrics} We measure the precision of SAR-Agent on both BIRD Mini-Dev and Spider 2.0-Snow.  
In addition, the Arcwise team reports 161 error examples in BIRD Mini-Dev through human audit \cite{BIRDminidev}.
We analyze the SAR-Agent hit rate on these examples, defined as the percentage of Arcwise detections that SAR-Agent also detects. Finally, we measure per-example monetary cost (USD) and the number of execution steps for SAR-Agent.



\minihead{Evaluation methods} Instead of outputting only a binary YES/NO decision when judging annotations, we collected the agent’s diagnostic report and manually verified the validity of its listed error reasons. We count an agent's detection as successful for a given text-to-SQL example if manual review confirms at least one of the error reasons it reports.

We further compared its diagnostic reports against the results of the Arcwise team's manual examination. Specifically, we counted: (1) examples flagged as incorrect by both SAR-Agent and the Arcwise team, (2) examples flagged only by SAR-Agent, and (3) examples flagged only by the Arcwise team.



\begin{table}[t]
\caption{Performance of SAR-Agent. On BIRD Mini-Dev, SAR-Agent achieves 83\% precision with an average cost of \$0.44 and 5.1 steps. On Spider 2.0-Snow, it achieves 89\% precision with an average cost of \$1.11 and 7.6 steps.}
\label{tab:agent_per}
    \centering
    \small
    \begin{tabular}{lcc}
        \toprule
        \textbf{Benchmarks} & \textbf{BIRD Mini-Dev} & 
        \textbf{Spider 2.0-Snow}\\
      \midrule
        \textbf{Precision (\%)} & 83 & 89\\
        \textbf{Average Cost (\$)} & 0.44 & 1.11 \\
        \textbf{Average Steps} & 5.1 & 7.6
        \\
    \bottomrule

    \end{tabular}
\end{table}

\begin{figure}[t]
    \centering
    \includegraphics[width=1\linewidth]{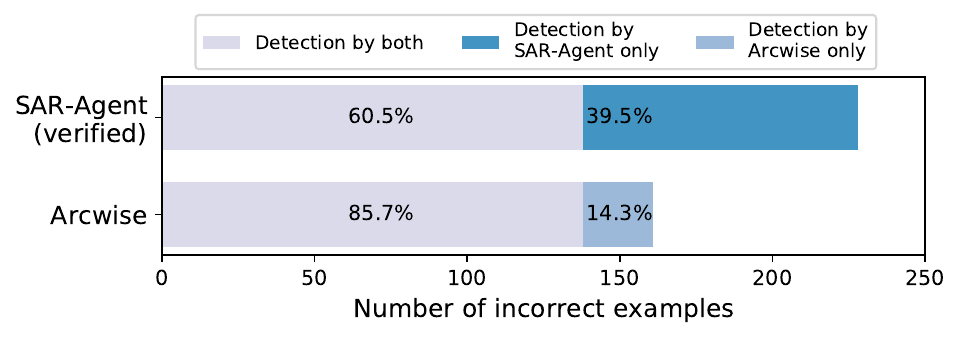}
    \caption{Number of detected error examples in BIRD Mini-Dev by SAR-Agent (after manual verification) and the Arcwise team. Of the 161 noisy examples identified by Arcwise, SAR-Agent achieves a hit rate of 85.7\%. Additionally, SAR-Agent identifies 41.6\% more error examples than Arcwise.}
    \label{fig:sar_comp}
\end{figure}

\subsection{Evaluation Results}

\minihead{Precision of SAR-Agent}
We evaluated SAR-Agent on BIRD Mini-Dev and Spider 2.0-Snow. We report the precision of SAR-Agent for both benchmarks in Table~\ref{tab:agent_per}. Among 498 examples in BIRD Mini-Dev, SAR-Agent flags 274 as incorrect. Our manual verification confirms 228 as true positive, yielding a precision of 83\%. On Spider 2.0-Snow, SAR-Agent flags 79 examples as incorrect. After manual verification, we find the provided error reasons are valid in 70 cases, yielding a precision of 89\%.

\minihead{Comparison of SAR-Agent with Arcwise SQL Experts}
We compare SAR-Agent against the SQL experts from the Arcwise team in Figure~\ref{fig:sar_comp}. After manual verification, we find that SAR-Agent successfully detects 138 of the 161 error examples reported by the Arcwise team, achieving an 85.7\% hit rate. Additionally, we compare the total number of error examples detected by SAR-Agent and the Arcwise team. SAR-Agent detects 41.6\% more error examples than the Arcwise team.

\minihead{Cost and execution steps of SAR-Agent} We report SAR-Agent’s monetary cost and the number of execution steps per example in Table~\ref{tab:agent_per}. On BIRD Mini-Dev, SAR-Agent spends \$0.44 per example with an average of 5.1 steps. On Spider 2.0-Snow, which features more complex queries and schemas, SAR-Agent requires more annotation-diagnosis steps, increasing the per-example cost to \$1.11 and the average number of steps to 7.6.

\begin{figure}[t]
    \centering
    \begin{subfigure}[t]{\linewidth}
        \centering
        \includegraphics[width=0.75\linewidth]{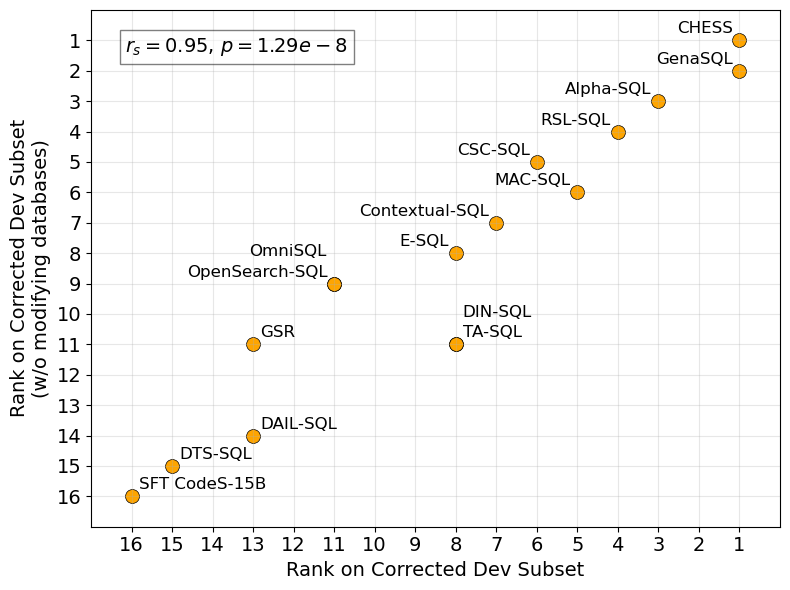}
        \caption{Correlation between EX rankings on the corrected Dev subset with and without database modification. $r_s = 0.95$ indicates a strong positive correlation.}
        \label{fig:aba_corr_1}
    \end{subfigure}%
    \vspace{1em}
    \begin{subfigure}[t]{\linewidth}
        \centering
        \includegraphics[width=0.75\linewidth]{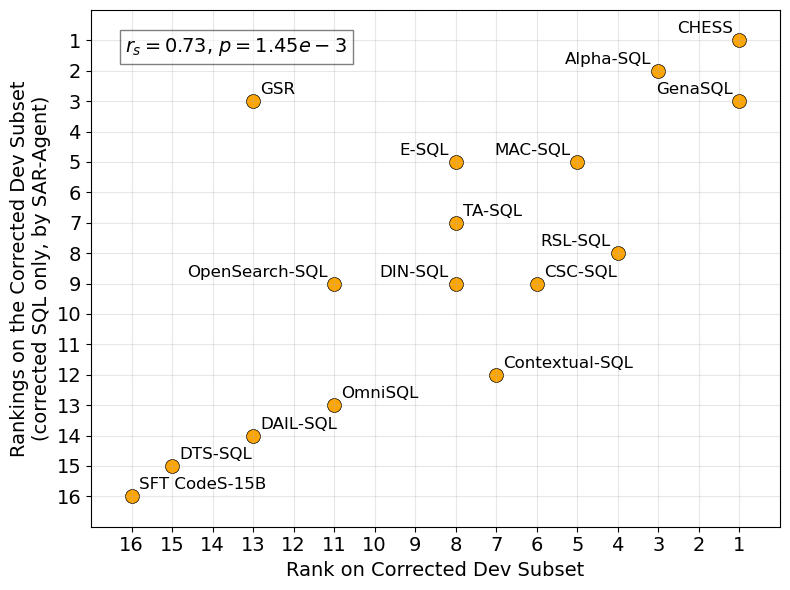}
        \caption{Correlation between EX rankings on the fully corrected Dev subset and the SQL-only-corrected Dev subset by SAR-Agent. $r_s = 0.73$ indicates strong correlation.}
        \label{fig:aba_corr_2}
    \end{subfigure}
    \vspace{1em}
    \begin{subfigure}[t]{\linewidth}
        \centering
        \includegraphics[width=0.75\linewidth]{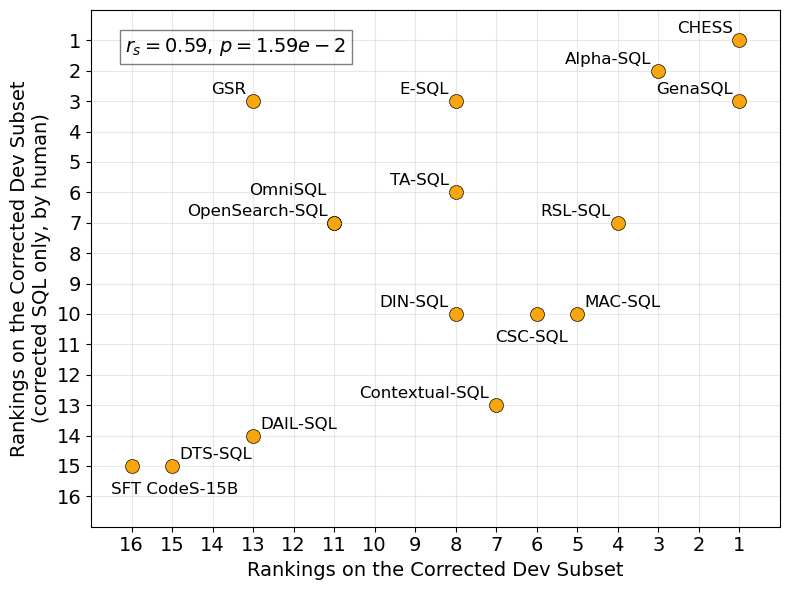}
        \caption{Correlation between EX rankings on the fully corrected Dev subset and the SQL-only-corrected Dev subset by humans. $r_s = 0.59$ indicates moderate correlation.}
        \label{fig:aba_corr_5}
    \end{subfigure}
    \caption{\blue{Rank correlation between the fully corrected Dev subset and corrected Dev subsets across settings.}}
    \label{fig:aba_corr}
\end{figure}
\section{ablation study}
\blue{We conducted three Spearman's rank correlation analyses to assess how different correction settings (i.e., (i) no database modification, (ii) SQL-only-corrected Dev subset by SAR-Agent, and (iii) SQL-only-corrected Dev subset by humans) affect agent rankings. For each setting, we compare EX rankings against those on the fully corrected Dev subset (Table \ref{tab:fixed_stat}). We describe the details of each correction setting in Appendix \ref{subsec:correction_setting}.}

\blue{We first compared EX rankings on the corrected Dev subset with and without database modification. As shown in Figure \ref{fig:aba_corr_1}, $r_s = 0.95$ indicates that database modification has minimal impact on the rankings of agents.}

\blue{We next evaluated SQL-only corrections, considering fixes produced by SAR-Agent and by humans. As shown in Figure \ref{fig:aba_corr_2} and Figure \ref{fig:aba_corr_5}, the correlation between EX rankings on the fully corrected Dev subset and the SQL-only-corrected Dev subset by SAR-Agent ($r_s = 0.73$) is stronger than that between the fully corrected Dev subset and the SQL-only-corrected subset by humans ($r_s = 0.59$). We attribute this gap to residual issues in the natural language input (e.g., ambiguity or domain-knowledge errors). SAR-Agent’s fixes assume these issues are resolved, making its SQL-only corrections closer to the fully corrected Dev subset. In contrast, human-repaired SQL queries align with the question and external knowledge as written, even when they contain issues.}

\section{Related work}

\minihead{Text-to-SQL benchmarks}
Researchers have proposed a wide range of datasets and benchmarks to evaluate the effectiveness of text-to-SQL approaches. Early research primarily uses single-domain datasets \cite{IMDB, scholar, ATIS}. In recent years, attention has shifted toward cross-domain datasets to better assess models’ generalization across diverse database schemas \cite{lee2021kaggledbqarealisticevaluationtexttosql, spider1, bird, wikisql}. The rapid progress in LLMs and text-to-SQL techniques has further driven the development of new benchmarks, which feature more complex SQL queries and database schemas \cite{spider2, BEAVER, livesqlbench2025, swesql, birdinteract2025}. In this work, we systematically analyze errors in two widely studied text-to-SQL benchmarks, BIRD Mini-Dev \cite{bird} and Spider 2.0-Snow \cite{spider2}. Our analysis reveals error rates exceeding 50\% across these benchmarks.

\minihead{Text-to-SQL methods}
Text-to-SQL approaches have a long history, with research evolving significantly in recent years thanks to advances in LLMs. Recent work has focused on developing text-to-SQL agents \cite{pourreza2023dinsqldecomposedincontextlearning, chess, opensearchsql, li2025alphasqlzeroshottexttosqlusing, wang2025macsqlmultiagentcollaborativeframework, gsrsql, genasql, agrawal2025text2sql, cscsql, cao2024rsl, esql,qu-etal-2024-generation, pourreza2024dtssqldecomposedtexttosqlsmall,gao2023texttosqlempoweredlargelanguage, ReFoRCE} that decompose the task into several sub-tasks and modules, including schema linking, SQL generation, and SQL refinement. Another major direction is fine-tuning LLMs for text-to-SQL tasks. Early efforts primarily relies on supervised learning \cite{XiYan, omnisql}, whereas recent approaches increasingly leverage reinforcement learning frameworks to strengthen the reasoning capabilities of LLMs \cite{cscsql, pourreza2025reasoningsqlreinforcementlearningsql}. In this work, we re-evaluated all open-source agents on the BIRD leaderboard using both the original and corrected versions of the BIRD Dev subset. The re-evaluation reveals significant changes in execution accuracy and agent rankings, demonstrating that annotation errors undermine the leaderboard’s reliability.

\minihead{Annotation errors in benchmarks}
\blue{Many studies have highlighted significant issues with existing benchmarks \cite{BIRDminidev, wretblad2024understandingeffectsnoisetexttosql, pourreza2023evaluatingcrossdomaintexttosqlmodels, NL2SQL-Empirical, abc, kapoor2024aiagentsmatter, yang2025agentoccamsimplestrongbaseline, aleithan2024swebenchenhancedcodingbenchmark, 10992485, liu2023codegeneratedchatgptreally, yu2025utboostrigorousevaluationcoding}.} In particular, several works have examined the presence of annotation errors in widely-used text-to-SQL benchmarks such as Spider and BIRD \cite{BIRDminidev, wretblad2024understandingeffectsnoisetexttosql, pourreza2023evaluatingcrossdomaintexttosqlmodels, NL2SQL-Empirical}. \blue{To fix errors in agentic benchmarks, prior works use the combination of automatic correction and manual verification. For example, in coding and software engineering benchmarks \cite{10992485, liu2023codegeneratedchatgptreally, yu2025utboostrigorousevaluationcoding}, researchers employ automatically generated test cases and oracles to strengthen the rigor of evaluating code generation. Moreover, manual verification and checks are used to ensure the soundness of error corrections \cite{aleithan2024swebenchenhancedcodingbenchmark, 10992485, liu2023codegeneratedchatgptreally, yu2025utboostrigorousevaluationcoding, abc}.}
\blue{Our work combines agent-powered automation with manual verification to audit and correct text-to-SQL benchmarks. We ran SAR-Agent to automatically detect and diagnose text-to-SQL annotation errors, with every reported error subsequently reviewed by SQL experts. Final adjudication on whether an annotation is incorrect was made by SQL experts. We corrected a sampled BIRD Dev subset based on SAPAR. In SAPAR, human annotators can adopt the SQL revision proposed by SAR-Agent after manual verification. If they disagree, they revise the SQL themselves. Annotators also need to revise the user question and supporting evidence when these contain errors.}



\minihead{LLM-as-a-Judge} LLM-as-a-Judge denotes using LLMs, which are typically aligned with human preferences via reinforcement learning from human feedback (RLHF), to evaluate system performance \cite{zheng2023judgingllmasajudgemtbenchchatbot}. As LLMs' capabilities, particularly reasoning, have improved, LLM-as-a-Judge has been integrated into evaluation workflows \cite{gu2025surveyllmasajudge} across finance \cite{brief2024mixingupcocktaileffect, yu2024finconsynthesizedllmmultiagent, wang2024quantagentseekingholygrail}, law \cite{ma2025leveraginglargelanguagemodels, ryu-etal-2023-retrieval}, and science \cite{krolik2024leveraginglargelanguagemodels, brake-schaaf-2024-comparing}. Recent work extends this paradigm to multimodal scenarios with multimodal large language models (MLLMs), termed MLLM-as-a-Judge \cite{chen2024mllmasajudgeassessingmultimodalllmasajudge}, and to agentic frameworks for evaluating agentic systems, termed Agent-as-a-Judge \cite{zhuge2024agentasajudgeevaluateagentsagents}. Because Text-to-SQL annotation requires not only database expertise but also domain knowledge across fields such as science, sports, and finance, we develop the first AI agent (SAR-Agent) to serve as a co-reviewer for text-to-SQL annotation. We hope SAR-Agent will facilitate the development of high-quality text-to-SQL benchmarks.

\section{Conclusion}

\blue{In this work, we quantify annotation error rates for two widely used text-to-SQL benchmarks, BIRD and Spider 2.0-Snow. We then correct a subset of the BIRD Dev set to systematically evaluate how annotation errors affect agent performance and leaderboard rankings. To facilitate error detection, we develop SAR-Agent, which generates a diagnostic report for each text-to-SQL annotation. After SQL experts manually verify each error reason listed in the report, we find error rates of 52.8\% and 62.8\% in BIRD Mini-Dev and Spider 2.0-Snow, respectively. We introduce SAPAR, which incorporates SAR-Agent into the existing text-to-SQL annotation pipeline, to correct a sampled BIRD Dev set. Re-evaluating all 16 open-source text-to-SQL agents from the BIRD leaderboard, we find relative performance changes from -7\% to +31\% and ranking shifts from -9 to +9 positions, underscoring the unreliability of current text-to-SQL leaderboards. We advocate that future work use SAR-Agent and SAPAR to develop high-quality text-to-SQL benchmarks.
}

\section{Acknowledgements}
We are grateful to the CloudLab \cite{cloudlab} for providing computing resources for experiments. This research was supported in part by Google and the Open Philanthropy project. 

\clearpage

\balance
\bibliographystyle{ACM-Reference-Format}
\bibliography{sample}

\clearpage
\nobalance

\appendix
\section{Appendix}

\begin{figure}[!t]
  \begin{subfigure}[t]{\columnwidth}
   \input{figures/bird416}
\caption{Input example for SAR-Agent (BIRD Mini-Dev example 416).}
\label{sub:probelm_input}
\end{subfigure}

 \vspace{1em}

  \begin{subfigure}[t]{\columnwidth}
    \input{figures/bird416_report}
\caption{Diagnostic report for BIRD Mini-Dev example 416 generated by SAR-Agent.}
\label{sub:probelm_output}
\end{subfigure}

\caption{Input and output of SAR-Agent on BIRD Mini-Dev example 416.}
\label{fig:fixed_sql_416}
\end{figure}
\begin{figure}[t]
\input{figures/bird416_agent}

\caption{Trajectory of SAR-Agent on BIRD Mini-Dev example 416.}
\label{fig:agent_tra}
\end{figure}
\subsection{Example}
\label{subsec:end_example}
\blue{We present an end-to-end example illustrating how SAR-Agent verifies an annotation and generates a diagnostic report. Using Example 416 from the BIRD Mini-Dev set as an example (Figure \ref{sub:probelm_input}), SAR-Agent takes as inputs the annotation (the user query, relevant external knowledge, and the ground-truth SQL query) along with the database schema. The resulting diagnostic report is shown in Figure \ref{sub:probelm_output}. In addition, we show the agent's reasoning trajectory in Figure \ref{fig:agent_tra}.}

\subsection{Configurations of Text-to-SQL Agents}
\label{subsec:config_agent}
We characterize the evaluated agents and their configurations in Table~\ref{tab:agents} based on the following six common modules.
\begin{enumerate}[leftmargin=*]
    \item \textit{Fine-tuning}, including supervised fine-tuning (SFT) and reinforcement learning (RL), enhances LLMs' ability to generate SQL queries.
    
    \item \textit{Question preprocessing} decomposes complex questions into simpler sub-questions or enriches user questions with the related database information. 
    
    \item \textit{Schema linking} preprocesses the database schema for a user question. Techniques differ in their schema representations, in how they map the question to that representation, and in how they retrieve database values.
   
    \item \textit{SQL generation} produces candidate SQL queries using zero-shot and few-shot prompting; the latter augments the prompt with a few in-context examples.
    
    \item \textit{SQL refiner} revises generated queries based on execution feedback from the database to fix issues such as syntax errors or empty results.
    
    \item \textit{SQL selection} selects among multiple generated SQL candidates. The most common method is self-consistency, which chooses the query with the most agreement across sampled outputs. Other approaches include training a classifier for selection and generating unit-test queries for selection.
    
\end{enumerate}

\begin{table*}[t]
\caption{Overview of the 16 open-source text-to-SQL agents from the BIRD leaderboard, including their respective models and modules. Abbreviations: RL = Reinforcement Learning; SFT = Supervised Fine-tuning.}
    \label{tab:agents}
    \centering
    \footnotesize
    \renewcommand{\arraystretch}{1.5} 
      \begin{threeparttable}
    \begin{tabular}{clcccccc}
        \toprule
        \multirow{2}{*}{\textbf{Agents}} & \multirow{2}{*}{\textbf{Models}} & \multicolumn{6}{c}{\textbf{Modules}} \\

        \cline{3-8}
        
        & & \textbf{Fine-tuning}  & \makecell*[c]{\textbf{Question} \\ \textbf{Preprocessing}} 
        & \makecell*[c]{\textbf{Schema}\\\textbf{Linking}}
        & \makecell*[c]{\textbf{SQL} \\ \textbf{Generation}}
        & \makecell*[c]{\textbf{SQL} \\ \textbf{Refiner}}
        & \makecell*[c]{\textbf{SQL}  \\ \textbf{Selection}} \\

        \midrule

         Contextual-SQL~\cite{agrawal2025text2sql} &
         \makecell*[l]{\textit{Generator model:} \\ Qwen2.5-Coder-32B-Instruct~\cite{qwen2025qwen25technicalreport} \\ \textit{Reward model:}\\ ctx-bird-reward-250121} &
         \mycross & \mycross & \mycross  & few-shot & \mycross & model-based \\ \hline
         
         CSC-SQL~\cite{cscsql} &
         \makecell*[l]{\textit{Merge model:} \\ Qwen2.5-Coder-7B-Instruct~\cite{qwen2025qwen25technicalreport} \\ \textit{Generation model:} \\ XiYanSQL-QwenCoder-32B-2412~\cite{XiYan}} &
         RL, SFT\tnote{1} & \mycross & \mycross & zero-shot & \mycheck & self-consistency  \\ \hline
         
         GenaSQL~\cite{genasql} &
         \makecell*[l]{text-embedding-3-small~\cite{textembedding}, \\ GPT-4o-2024-08-06~\cite{openai2024gpt4ocard}\tnote{2}} &
         \mycross & \mycross & \mycheck  & few-shot & \mycheck &
         \makecell*[c]{self-consistency, \\ model-based} \\ \hline
         
         OpenSearch-SQL~\cite{opensearchsql} &
         \makecell*[l]{\textit{Retrieval model:} bge-m3~\cite{bge} \\ \textit{Generation model:} \\ GPT-4o-2024-05-13~\cite{openai2024gpt4ocard}} &
         \mycross & \mycross & \mycheck  & few-shot & \mycheck & self-consistency \\ \hline
         
         OmniSQL~\cite{omnisql}\tnote{3} & 
         OmniSQL-32B & SFT & \mycross & \mycross & zero-shot & \mycross & single candidate
         \\ \hline
         
         CHESS~\cite{chess}\tnote{4} &
         GPT-4o-2024-08-06~\cite{openai2024gpt4ocard} & \mycross & \mycross & \mycheck  & zero-shot & \mycheck & test-based \\ \hline
         
         Alpha-SQL~\cite{li2025alphasqlzeroshottexttosqlusing} &
         \makecell*[l]{\textit{Retrieval model:} \\ text-embedding-3-large~\cite{textembedding} \\ \textit{Generation model:} \\ Qwen2.5-Coder-32B-Instruct~\cite{qwen2025qwen25technicalreport}} &
         \mycross & \mycheck & \mycheck & \makecell*[c]{zero-shot} & \mycheck & self-consistency \\ \hline
         
         GSR~\cite{gsrsql} &
         GPT-4o-2024-11-20~\cite{openai2024gpt4ocard} & \mycross & \mycheck & \mycheck & zero-shot & \mycheck & single candidate \\ \hline
         
         RSL-SQL~\cite{cao2024rsl} &
         GPT-4o-2024-08-06~\cite{openai2024gpt4ocard} & \mycross & \mycross & \mycheck & few-shot & \mycheck & model-based \\ \hline
         
         E-SQL~\cite{esql} &
         GPT-4o-2024-08-06~\cite{openai2024gpt4ocard} & \mycross & \mycheck & \mycheck & few-shot & \mycheck & single candidate \\ \hline
         
         TA-SQL~\cite{qu-etal-2024-generation}\tnote{5} &
         GPT-4o-2024-08-06~\cite{openai2024gpt4ocard} & \mycross & \mycross & \mycheck & few-shot & \mycross & single candidate \\ \hline
         
         DTS-SQL~\cite{pourreza2024dtssqldecomposedtexttosqlsmall} &
         DeepSeek 7B~\cite{deepseek-llm} & SFT & \mycross & \mycheck & zero-shot & \mycross & single candidate \\ \hline
         
         MAC-SQL~\cite{wang2025macsqlmultiagentcollaborativeframework}\tnote{5} &
         GPT-4o-2024-08-06~\cite{openai2024gpt4ocard} & \mycross & \mycheck & \mycheck & few-shot & \mycheck & single candidate \\ \hline
         
         SFT CodeS-15B~\cite{codes} &
         SFT CodeS-15B & SFT & \mycross & \mycheck & zero-shot & \mycross & single candidate \\ \hline
         
         DAIL-SQL~\cite{gao2023texttosqlempoweredlargelanguage}\tnote{5} &
         GPT-4o-2024-08-06~\cite{openai2024gpt4ocard} & \mycross & \mycross & \mycheck & few-shot & \mycross & self-consistency \\ \hline
         
         DIN-SQL~\cite{pourreza2023dinsqldecomposedincontextlearning}\tnote{5} &
         GPT-4o-2024-08-06~\cite{openai2024gpt4ocard} & \mycross & \mycheck & \mycheck & few-shot & \mycheck & single candidate \\
 
         \bottomrule

    \end{tabular}
    \begin{tablenotes}
    \item[1] Although CSC-SQL does not incorporate the SFT process, its SQL generation model, XiYanSQL-QwenCoder-32B-2412, is a fine-tuned model for text-to-SQL tasks.
    \item[2] Since we do not have access to AWS Bedrock or Google Cloud, we use OpenAI text-embedding-3-small as our embedding model and GPT-4o-2024-08-06 for the other modules in GenaSQL.
    \item[3] We evaluate OmniSQL under the greedy decoding strategy.
    \item[4] We configure CHESS with three modules (Information Retriever, Candidate Generator, and Unit Tester) since this setup ranks highest on the BIRD leaderboard.
    Since we do not have access to Google Cloud, we run CHESS with GPT‑4o instead.
    \item[5] Due to budget constraints, we run TA-SQL, MAC-SQL, DAIL-SQL, and DIN-SQL with GPT-4o instead of GPT-4.
  \end{tablenotes}
  \end{threeparttable}
\end{table*}

\subsection{Correction Settings on the BIRD Dev Subset }
\label{subsec:correction_setting}
\blue{We consider four correction settings using a randomly sampled 100-example subset of the BIRD Dev set. The settings are:}
\begin{enumerate}[leftmargin=*]
    \item \blue{\textit{Fully corrected Dev subset:} We corrected issues in the user question, external knowledge, ground-truth SQL, and database schema, and modified the database when necessary. The statistics for these corrections are shown in Table~\ref{tab:fixed_stat}.}
    
    \item \blue{\textit{Corrected Dev subset without database modification:} We corrected issues in  the question, external knowledge, ground-truth SQL, and database schema, but left the database unchanged.}
    
    \item \blue{\textit{SQL-only-corrected Dev subset by SAR-Agent:} We replaced only the misannotated ground-truth SQL with the repair generated by SAR-Agent. SAR-Agent produces fixed queries under the assumption that the user question and external knowledge have also been fixed. In this setting, however, the user question and external knowledge are left unchanged.}
   
    \item \blue{\textit{SQL-only-corrected Dev subset by humans:} We replaced only the misannotated ground-truth SQL with a human-repaired query.  In this setting, we left the user question and external knowledge unchanged and aligned the repaired SQL with them, even if they contained errors.}
\end{enumerate}

\blue{We evaluated all 16 open-source text-to-SQL agents on the original (uncorrected) BIRD Dev subset and on four corrected Dev subsets. We report each agent’s execution accuracy and rankings on every subset in Table~\ref{tab:rerank}.
CHESS~\cite{chess}, which ranks 7th on the original Dev subset, consistently ranks 1st across all four corrected subsets.}

\begin{table*}[t]
\caption{\blue{Execution accuracy and rankings of 16 open-source text-to-SQL agents on the original Dev subset (uncorrected) and four corrected Dev subsets.}}
    \label{tab:rerank}
    \centering
    \small
    \renewcommand{\arraystretch}{1.5} 
      \begin{threeparttable}
    \begin{tabular}{c rr rr  rr  rr  @{\hspace{1em}} rr @{\hspace{1.5em}}}
        \toprule
        \multirow{2}{*}{\textbf{Agents}} & \multicolumn{2}{c}{\makecell*[c]{\textbf{Original}\\\textbf{Dev subset}}} 
        & \multicolumn{2}{c}{\makecell*[c]{\textbf{Fully corrected}\\\textbf{ Dev subset}}} 
        & \multicolumn{2}{c@{\hspace*{-1.5em}}}{\makecell*[c]{\textbf{Corrected Dev subset w/o}\\ \textbf{ database modification}}} 
        & \multicolumn{2}{@{\hspace*{3em}}c@{\hspace*{-0.5em}}}{\makecell*[c]{\textbf{SQL-only-corrected Dev}\\ \textbf{subset by SAR-Agent}}} 
        & \multicolumn{2}{@{\hspace*{1.5em}}c@{\hspace*{0.5em}}}{\makecell*[c]{\textbf{SQL-only-corrected Dev}\\ \textbf{subset by humans}}}\\

        \cline{2-11}
        
       & \textbf{EX}  & \textbf{Ranking} & 

        \textbf{\hspace{1em} EX}  & \textbf{Ranking} & 

        \textbf{\hspace{2em} EX}  & \textbf{\hspace*{-2em} Ranking}& 

        \textbf{\hspace{3.5em} EX}  & \textbf{\hspace{-2em} Ranking}& 

        \textbf{\hspace{3em} EX}  & \textbf{Ranking} \\

        \midrule

         Contextual-SQL &
         \textbf{73} & \textbf{1} &
         71 & 7 &
         \hspace{2em}74 & 7 &
         63 & 12 &
         69 & 13\\

         OpenSearch-SQL &
         71 & 2 &
         69 & 11 &
         \hspace{2em}71 & 9 &
         64 & 9 &
         71 & 7\\ 

         CSC-SQL &
         67 & 3 &
         74 & 6 &
         \hspace{2em}76 & 5 &
         64 & 9 &
         70 & 10\\ 

         GSR &
         66 & 4 &
         68 & 13 &
         \hspace{2em}70 & 11 &
         69 & 3 &
         75 & 3\\ 

         Alpha-SQL &
         65 & 5 &
         78 & 3 &
         \hspace{2em}80 & 3 &
         71 & 2 &
         76 & 2\\ 

         GenaSQL &
         64 & 6 &
         \textbf{81} & \textbf{1} &
         \hspace{2em}81 & 2 &
         69 & 3 &
         75 & 3\\ 

         CHESS &
         62 & 7 &
         \textbf{81} & \textbf{1} &
         \hspace{2em}\textbf{83} & \textbf{1} &
         \textbf{76} & \textbf{1} &
         \textbf{80} & \textbf{1}\\ 

         E-SQL &
         62 & 7 &
         70 & 8 &
         \hspace{2em}73 & 8 &
         68 & 5 &
         75 & 3\\ 

         OmniSQL &
         60 & 9 &
         69 & 11 &
         \hspace{2em}71 & 9 &
         62 & 13 &
         71 & 7 \\ 

         RSL-SQL &
         60 & 9 &
         76 & 4 &
         \hspace{2em}78 & 4 &
         66 & 8 &
         71 & 7\\ 

         DTS-SQL &
         60 & 9 &
         56 & 15 &
         \hspace{2em}57 & 15 &
         53 & 15 &
         58 & 15\\ 

         TA-SQL &
         59 & 12 &
         70 & 8 &
         \hspace{2em}70 & 11 &
         67 & 7 &
         74 & 6\\ 

         DIN-SQL &
         59 & 12 &
         70 & 8 &
         \hspace{2em}70 & 11 &
         64 & 9 &
         70 & 10\\ 

         MAC-SQL &
         58 & 14 &
         75 & 5 &
         \hspace{2em}75 & 6 &
         68 & 5 &
         70 & 10\\ 

         DAIL-SQL &
         56 & 15 &
         68 & 13 &
         \hspace{2em}67 & 14 &
         58 & 14 &
         63 & 14\\ 

         SFT CodeS-15B &
         56 & 15 &
         52 & 16 &
         \hspace{2em}53 & 16 &
         52 & 16 &
         58 & 15\\ 

         \bottomrule

    \end{tabular}
   
  \end{threeparttable}
\end{table*}

\begin{figure}[t]
\input{figures/sf_local015}

\caption{Example of an additional error after updating Spider 2.0-Snow problem sf\_local015 by the Spider 2.0 team. The added formula calculates the number of fatalities per collision rather than the percentage of motorcyclist fatalities.}
\label{fig:spider_fix}
\end{figure}

\subsection{Case study}
The Spider 2.0 team updated their user questions to resolve ambiguity on July 13, 2025. In this section, we study whether their fix addressed annotation errors.

\minihead{Data settings} Among 121 examples with released gold queries, the Spider 2.0 team updated user questions in 71 examples. The updated set is called Spider 2.0-Snow, while the original set is called Spider 2.0-Snow-0713.\footnote{Spider 2.0-Snow can be found at: \url{https://github.com/xlang-ai/Spider2/blob/main/spider2-snow/spider2-snow.jsonl}, while Spider 2.0-Snow-0713 can be found at: \url{https://github.com/xlang-ai/Spider2/blob/main/spider2-snow/spider2-snow-0713.jsonl}} 

\minihead{Evaluation method} We ran SAR-Agent on Spider 2.0-Snow-0713 to produce a diagnostic report for each example and manually verified the reported error reasons. 

\minihead{Evaluation results} 
We compare error rates for the old and updated versions of Spider 2.0-Snow. We find their correction reduces the error rate from 66.1\% to 62.8\%, implying that errors in only 3.3\% of examples are fully resolved. This modest reduction indicates that the update by Spider 2.0 team has limited effectiveness.

We also find that their fix introduces additional errors in 6 examples.\footnote{Because these examples also contain other errors detected previously, the overall error rate of Spider 2.0-Snow is unchanged.} For example, as illustrated in Figure ~\ref{fig:spider_fix}, they added a formula that computes the number of fatalities per collision in the user question. However, this formula conflicts with the term ``the percentage of motorcyclist fatalities'' in the question. If some collisions result in more than one fatality, the resulting percentage of motorcyclist fatalities could exceed 100\%, which is invalid. Moreover, because the corresponding gold SQL query multiplies the number of motorcyclist fatalities per collision by 100, even revising the question to ``the number of motorcyclist fatalities per collision'' would not resolve the error in this example. Therefore, only updating user questions is insufficient to fix the annotation errors in Spider 2.0-Snow. We advocate that future work refine the entire Spider 2.0-Snow benchmark based on our proposed SAPAR to improve annotation quality.

\end{document}